\documentclass[journal,12pt]{elsarticle}

\usepackage{lineno}
\usepackage{amsmath,amsfonts}
\usepackage{array}
\usepackage[caption=false,font=normalsize,labelfont=sf,textfont=sf]{subfig}
\usepackage{textcomp}
\usepackage{stfloats}
\usepackage{url}
\usepackage{verbatim}
\usepackage{graphicx}
\usepackage[autostyle=true]{csquotes} 
\usepackage{hyperref}
\usepackage[linesnumbered,noline,ruled,commentsnumbered]{algorithm2e}
\usepackage{bm}
\usepackage{enumitem}
\usepackage{booktabs}
\usepackage{multirow}
\usepackage{array}
\usepackage{caption}
\usepackage{makecell}
\usepackage{xurl}

\usepackage{xcolor,colortbl}
\definecolor{migris}{rgb}{0.78,0.78,0.78}
\newcommand{\mejor}{\cellcolor{migris}}  
\definecolor{migris_2}{rgb}{0.92,0.92,0.92}

\usepackage{rotating}
\newcommand{\PreserveBackslash}[1]{\let\temp=\\#1\let\\=\temp}
\newcolumntype{C}[1]{>{\PreserveBackslash\centering}p{#1}}
\newcolumntype{R}[1]{>{\PreserveBackslash\raggedleft}p{#1}}
\newcolumntype{L}[1]{>{\PreserveBackslash\raggedright}p{#1}}
\usepackage{adjustbox}

\begin{document}

\begin{frontmatter}



\title{On the Inherent Robustness of One-Stage Object Detection against Out-of-Distribution Data}




\author[address0]{Aitor Martinez-Seras\corref{mycorrespondingauthor1}}
\cortext[mycorrespondingauthor1]{Corresponding author 1: Arriaga Kalea, 2, 20870 Elgoibar, Gipuzkoa, Spain. Phone: +34943748000.}
\ead{amartinezseras@ideko.es}
\author[address1,address2]{Javier~Del~Ser\corref{mycorrespondingauthor2}}
\cortext[mycorrespondingauthor2]{Corresponding author 2: Parque Tecnologico de Bizkaia, 700, 48160 Derio, Bizkaia, Spain. Phone: +34946430850.}
\ead{javier.delser@tecnalia.com}
\author[address2]{Aitzol~Olivares-Rad}
\author[address1,address3]{Alain~Andres}
\author[address3]{Pablo~Garcia-Bringas}

\address[address0]{IDEKO, Basque Research and Technology Alliance (BRTA), 20870 Elgoibar, Spain}
\address[address1]{TECNALIA, Basque Research and Technology Alliance (BRTA), 48160 Derio, Spain}
\address[address2]{University of the Basque Country (UPV/EHU), 48940 Leioa, Spain}
\address[address3]{University of Deusto, 48007 Bilbao, Spain}

\begin{abstract}

Robustness is a fundamental aspect for developing safe and trustworthy models, particularly when they are deployed in the open world. In this work we analyze the inherent capability of one-stage object detectors to robustly operate in the presence of out-of-distribution (OoD) data. Specifically, we propose a novel detection algorithm for detecting unknown objects in image data, which leverages the features extracted by the model from each sample. Differently from other recent approaches in the literature, our proposal does not require retraining the object detector, thereby allowing for the use of pretrained models. Our proposed OoD detector exploits the application of supervised dimensionality reduction techniques to mitigate the effects of the curse of dimensionality on the features extracted by the model. Furthermore, it utilizes high-resolution feature maps to identify potential unknown objects in an unsupervised fashion. Our experiments analyze the Pareto trade-off between the performance detecting known and unknown objects resulting from different algorithmic configurations and inference confidence thresholds. We also compare the performance of our proposed algorithm to that of logits-based post-hoc OoD methods, as well as possible fusion strategies. Finally, we discuss on the competitiveness of all tested methods against state-of-the-art OoD  approaches for object detection models over the recently published Unknown Object Detection benchmark. The obtained results verify that the performance of avant-garde post-hoc OoD detectors can be further improved when combined with our proposed algorithm.
\end{abstract}


\end{frontmatter}


\section{Introduction}

The rapid advancement and widespread adoption of Artificial Intelligence (AI) systems for real-world applications have underscored the urgent need for these models to be safe and trustworthy \cite{bengio2024managing}. What trustworthiness means for AI is a highly debated concern in the recent years, attracting significant interest from the research community. A major breakthrough in shedding light on trustworthy AI was the publication of the \enquote{Ethical Guidelines for Trustworthy AI} \cite{ai_high-level_2019} in 2019 by the European Union, a regulatory actor of relevance in this matter. The text outlines seven key requirements for trustworthy AI and identify three essential pillars for fulfilling these requirements. The  \emph{technical robustness and safety} of an AI system is recognized as one of the seven requirements and a fundamental pillar of trustworthiness \cite{diaz-rodriguez_connecting_2023}. More recently, another important actor in the regulatory landscape, the National Institute of Standards and Technology (NIST) has defined in \cite{tabassi_artificial_2023} the technical robustness as a system's capacity to sustain its performance across diverse conditions. It entails not only consistent functioning under anticipated scenarios, but also the ability to minimize potential risks to individuals when operating in environments subject to unexpected events.

For this purpose, ensuring the safety of Machine Learning (ML) models involves developing robust systems capable of handling unknown semantics, effectively distinguishing between known and unknown data instances. When it comes to object detection tasks from image data, these models must be able to navigate open-world environments by discerning between background (irrelevant information), known objects (relevant information within the training distribution), and unknown objects (relevant information outside the training distribution). In this context, substantial efforts have been directed towards the field of Open World Object Detection (OWOD) \cite{joseph_towards_2021}\nocite{singh_order_2021,zhao_revisiting_2023}-\cite{wang_random_2023}. The goal pursued in this research area is to develop models capable of detecting unknown objects and incrementally learning new categories over time. Predominantly, research in this area has concentrated on methodologies that require retraining two-stage object detection models, such as Faster R-CNN. Different retraining strategies are designed to endow the learned model with the capability to discern a broader variety of objects (including those not present in the training data), such as continual learning approaches or loss functions related to the presence of objects (\emph{objectness}). Conversely, pretrained object detection models require less computational power when compared to retrained models because they are optimized on a fixed dataset of known classes. However, this efficiency comes at the cost of increased bias towards detecting only familiar objects, making them less suitable for open-world scenarios where new, unseen objects may appear. 

This manuscript contributes to understanding the robustness of pretrained object detection models in detecting unknown objects in image data. We depart from our hypothesis that, without retraining, single-stage object detection models can inherently detect unknown objects. To explore this, we introduce a simple OoD detection algorithm based on neural activations (\emph{feature maps}) of pretrained models. Additionally, we present an unsupervised learning method that leverages feature maps to enhance recall for unknown objects. Our extensive experimental setup evaluates the algorithm across different configurations and parameter choices, assessing its effectiveness in identifying known and unknown objects compared to logits-based OoD methods and retrained object detection models. We further explore potential improvements through a fusion of tested methods, inspired by findings in \cite{martinez2023novel}. Finally, we compare our approach with state-of-the-art OWOD methods, showing that the proposed algorithm achieves superior detection scores in the benchmark without requiring model retraining.

The remainder of the paper is organized as follows: Section \ref{sec:rel_work} revisits OoD detection techniques for object detection, along with a brief summary of the most influential works in OWOD and a detailed statement of the contribution brought by this work to the related literature. Section \ref{sec:Fmap_full_descr} presents and describes our algorithm for detecting unknown objects\footnote{Throughout this work \enquote{unknown object} and \enquote{OoD object} are used interchangeably.} in images, beginning with an overview of its compounding algorithmic steps, and followed by a detailed description of the method. This section also introduces two techniques aimed at enhancing the overall performance of the algorithm. The experimental setup is outlined in Section \ref{sec:exp_setup}, whereas the results are analyzed and discussed in Section \ref{sec:results}. Finally, Section \ref{sec:conclusions} concludes the manuscript with a summary of the main findings and an outlook towards potential research paths to be addressed in the future.

\section{Related Work and Contribution}\label{sec:rel_work}


This section provides an overview of existing OoD techniques tailored for object detection (Subsection \ref{ssec:rel_work_ood}), reviews key works within the OWOD framework (Subsection \ref{ssec:rel_work_OWOD}), and presents this paper's contributions to the OoD literature (Subsection \ref{ssec:contribution}).

\subsection{Out-of-Distribution Detection in Object Detection}\label{ssec:rel_work_ood}

As noted in the introduction, the goal of OoD detection is to identify unknown samples in classification tasks—specifically, samples with semantics not present in the training distribution. Numerous techniques have been proposed in this area.
e.g., ReAct \cite{sun_react_2021}, GradNorm \cite{huang_importance_2021}, MSP \cite{hendrycks_baseline_2016}, ODIN \cite{liang_enhancing_2017}, Energy \cite{liu_energy-based_2020} or Mahalanobis \cite{denouden2018improving}. These techniques were originally conceived for detecting OoD data in image classification tasks. In object detection tasks, however, OoD detection refers to the model's ability to locate and identify objects in an image that do not belong to any of the classes it was trained to recognize. While traditional object detection models are trained to detect a fixed set of classes, OoD detection enables these models to recognize when an object falls outside this known set, flagging it as \emph{unknown}.

Due to its higher complexity than in traditional image classifier, the detection of unknowns in object detection models has garnered increasing interest during the last couple of years. Several methods devised to cope with this problem have been reported in the literature, such as those presented in \cite{du_vos_2022} and \cite{du_unknown-aware_2022}. Both are focused on regularizing a two-stage object detection model via the usage of unknowns. In the former, Virtual Outlier
Synthesis (VOS) technique is proposed to synthesize virtual outliers in the feature space, without relying on external data to allow the model to better learn the boundaries between known and unknown classes. In the latter, the same objective is pursued. Differently, unknowns are obtained or distilled from videos leveraging contiguous frames and a dissimilarity metric based on the $L_2$ distance between features of the object proposals.


Another example is \cite{wilson2023safe}, where authors propose a post-hoc OoD detector for object detection that utilizes Sensitivity-Aware FEatures (SAFE) extracted from residual convolutional layers and abnormal batch normalization activations. SAFE extracts object-specific feature vectors and employs a multilayer perceptron trained on the surrogate task of distinguishing adversarially perturbed samples from clean samples to classify each detected object as in-distribution or OoD.
Finally, the work in \cite{zolfi2024yolood} does not directly perform OoD detection in an object detection task. Instead, they leverage concepts from object detection to perform OoD detection in multi-label classification tasks. They convert a regular object detection model (YOLO) into an image classifier with inherent OoD detection capabilities by exploiting the model's ability to distinguish between objects of interest and irrelevant objects.

\subsection{Open-World Object Detection Framework}\label{ssec:rel_work_OWOD}

The OWOD framework was introduced by \cite{joseph_towards_2021} to address the limitations of traditional object detection models that assume all object classes are known during training. In the open-world setting, a model is expected to not only detect known objects but also identify unknown objects that do not belong to the training classes. The evaluation protocol established was based on the concepts and metrics introduced in \cite{dhamija_overlooked_2020}. It consist of grouping the classes of into sequential tasks, where each task introduces new classes incrementally and the not-yet-included classes are treated as the unknowns to be detected. This protocol is based on the \texttt{PASCAL VOC} \cite{everingham_pascal_2010} and \texttt{COCO} 
\cite{lin_microsoft_2014}
datasets, which are two of the most commonly used benchmark datasets in object detection. All \texttt{PASCAL} classes and data are considered to be the first task. The remaining 60 classes of \texttt{COCO} are grouped into three successive tasks with semantic drifts that are incrementally introduced to simulate the unknowns arriving at the model's input. Since its inception, this evaluation protocol has been adopted by most contributions dealing with open-world object detection tasks.


In the literature related to OWOD, it is important to note that most methods proposed to date hinge on the addition of an \enquote{unknown} class to a two-stage object detection model. This newly added class is trained using pseudo-labels obtained by several heuristics or techniques. Thereupon, each contribution to the literature following this approach introduces different algorithmic updates to the training algorithm to improve the performance of the model over the \enquote{unknown} class, while preserving the detection capabilities on the known classes. 

The solution proposed in \cite{joseph_towards_2021} called ORE uses contrastive clustering, an unknown-aware proposal network and an energy-based unknown identification to address the challenges of open-world detection. This way ORE is able to predict unknown objects as \enquote{unknown} and incrementally learn new classes as their labels become available, without forgetting previously learned classes. Building on this work, \cite{singh_order_2021} proposed a method called ORDER for addressing the challenges of detecting both known and unknown objects in road scenes. The authors introduce Feature-Mix, which improves unknown object detection by mixing features of known and unknown objects at the feature level. Additionally, they propose a focal regression loss to address intra-class scale variation and improve small object detection. The model is further improved through curriculum learning and is evaluated on the BDD100K \cite{yu2020bdd100k} and IDD \cite{varma2019idd} road scene datasets.

Contrary to previous approaches, the work done in \cite{gupta_ow-detr_2022} introduces a novel end-to-end transformer-based framework for open-world object detection called OW-DETR. It comprises three dedicated components to address the challenges of OWOD, specifically attention-driven pseudo-labeling, novelty classification, and objectness scoring. The attention-driven pseudo-labeling scheme selects object query boxes with high attention scores as potential unknown objects, which are then used along with ground-truth known objects to train a novelty classifier. The objectness branch aims to effectively separate foreground objects (known and pseudo-unknown) from the background by enabling knowledge transfer from known to unknown classes. Finally, the novelty classification branch is in charge of distinguishing the pseudo unknowns from each of the known classes.

Authors in \cite{zhao_revisiting_2023} point out that while the prior OWOD work introduced the problem definition, the experimental settings were unreasonable, the metrics were incomplete and the methodology was inappropriate. To address these issues, the paper proposes five fundamental benchmark principles to guide the construction of OWOD benchmarks. 
Furthermore, the authors present a novel OWOD method that includes an auxiliary proposal advisor module based on a selective search algorithm \cite{uijlings_selective_2013} to assist the RPN in identifying accurate unknown proposals by the addition of unsupervised information, and a class-specific expelling classifier to calibrate overconfident predictions and to filter out confusing outputs.

The work in \cite{wu_uc-owod_2022} goes one step beyond previous OWOD approaches by not only aiming to detect unknown instances, but also by classifying them into different unknown classes. First, an unknown label-aware proposal module and an unknown-discriminative classifier head are used to detect known and unknown objects. Then, similarity-based unknown classification and unknown clustering refinement modules are constructed to distinguish among multiple unknown classes.

A common problem to above introduced approaches is that the module generating the proposals for the unknown objects (namely, the RPN in two-stage object detectors) is highly biased towards detecting known objects as it is the only ground truth available in the training data. To mitigate this bias, RandBox \cite{wang_random_2023} substitutes the RPN with random region proposals that prevents the training from being confounded by the limited known object classes. Additionally, the authors introduce a matching score that does not penalize random proposals whose predictions do not match the known classes, encouraging the exploration of unknown objects all over the image.

Finally, the authors in \cite{liang_unknown_2023} address exclusively the initial stage of the OWOD framework, focusing solely on the identification of unknown objects. They evaluate their method using their proposed Unknown Object Detection (UOD) benchmark, which is derived from two small subsets of the \texttt{COCO} dataset, augmented with annotations for unknown objects. These new annotations are generated by them and their team. The evaluation protocol involves assessing the models only on one task within these two subsets, without any incremental addition of new classes to the model. The authors also propose the so-called unknown sniffer (UnSniffer) model, which introduces a Generalized Object Confidence (GOC) score to avoid improper suppression of unknowns. Similarly, a negative energy suppression loss is used to further suppress non-object samples in the background. To obtain the best bounding box for unknown objects during inference, the paper presents a graph-based determination scheme to replace the non-maximum suppression (NMS) post-processing.

\subsection{Contribution}\label{ssec:contribution}

In the light of the reviewed OWOD literature, it becomes evident that there is a clear trend towards using two-stage models that undergo retraining to acquire the capability to detect unknown objects. To avoid this computational burden, it is imperative to analyze the performance of OoD techniques for one-stage models which do not require retraining of the object detection model. In our research, we align our approach with the work of \cite{liang_unknown_2023}, by focusing exclusively on the first task of OWOD (the ability to detect unknown objects), without subsequently retraining the model with these newly identified classes. Our contributions in this regard can be summarized as follows:
\begin{itemize}[leftmargin=*]
	\item \emph{We explore the effectiveness of classic post-hoc OoD techniques on one-stage object detectors}. Specifically, we investigate the performance of traditional post-hoc OoD methods when applied to one-stage object detection models, which are generally more efficient but less studied in this context.
	\item \emph{We propose a simple OoD technique based on feature maps that does not require retraining}. We introduce a straightforward and computationally efficient method for detecting unknown objects that leverages feature maps within one-stage detector object detection models. This technique is designed to operate without the need for retraining, hence having no impact on the overall training complexity and maintaining the naive capability of the model to detect and identify known objects.
    \item \emph{We propose techniques for enhancing detection of potential unknown objects using feature maps}. We experiment with various methods to improve the detection of unknown objects by utilizing feature maps. This includes refining feature extraction and leveraging advanced techniques to better identify and classify OoD objects.
    \item \emph{We examine the potential of fusing OoD methods to robustly identify unknown objects}. We explore the benefits of combining different OoD detection methods in the form of a detection ensemble, aiming to complement each other when deciding on the known or unknown nature of an object. Similarly to supervised learning ensembles, fusing OoD methods can yield more robust OoD detection capabilities for the overall model than every one of them in isolation. We also show experimentally that the fusion of OoD methods with the one proposed in this work performs competitively when compared to state-of-the-art object detectors that detect unknowns by relying on specialized training procedures over pseudo-labels.
    
    
\end{itemize}

Ultimately, our research work contributes to the understanding and development of more effective OoD detection techniques for one-stage object detectors, leading to more robust and adaptable models for object detection in open-world scenarios.

\section[Open-World Object Detection by Feature Map Characterization]{Open-World Object Detection by Feature Map Characterization}\label{sec:Fmap_full_descr}

In this section we introduce our proposed OoD detection algorithm which, given its reliance on the feature maps within the model, is hereafter referred to as the FMap detector. Subsection \ref{ssec:fmap_high_level} presents the general workflow of FMap, while Subsection \ref{ssec:fmap_low_level} details in depth each of its compounding algorithmic steps. Next, Subsection \ref{ssec:fmap_sdr} elaborates on the benefits of employing Supervised Dimensionality Reduction (SDR) with our detector. Finally, Subsection \ref{ssec:fmap_eul} outlines an approach to enhance the capabilities for identifying unknown objects without penalizing the performance in detecting known objects.

\subsection{High-level Description of FMap}\label{ssec:fmap_high_level}

\begin{figure*}[!h]
    \centering
    \includegraphics[width=1\textwidth]{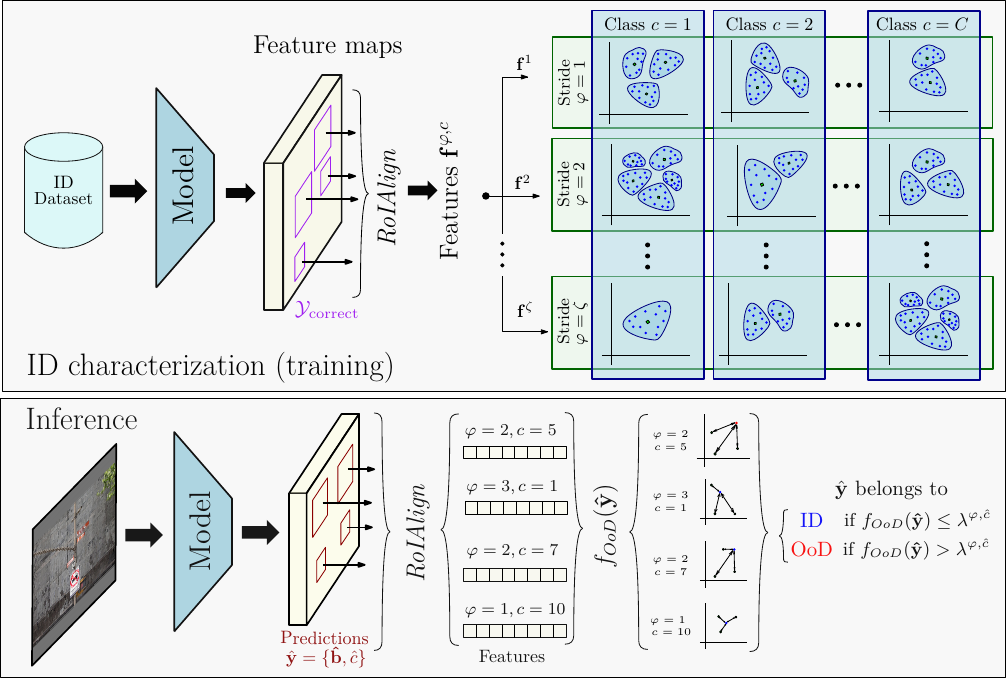}
    \caption[Description of the operation of the FMap detector]{Block diagram describing the general operation of the proposed FMap detector. In the inference part, an example for the stride ($\varphi$) and class ($c$) values is displayed.}
    \label{fig:fmap_detector}
\end{figure*}

We begin by pausing at the general workflow followed by FMap, which is illustrated in Figure \ref{fig:fmap_detector}. In essence, FMap characterizes the activation space at the output of a single-stage object detection model, i.e., we characterize the feature maps computed by the model for its input. In doing so, it is important to first note that modern single-stage object detection models employ a Feature Pyramid Network (FPN) \cite{lin_feature_2017} to enhance their ability to detect objects at various scales. Therefore, the characterization of activations made by FMap must account for this feature. In the case of two-stage detectors, feature maps of different strides are merged and standardized to have a uniform number of output channels or dimensions after the FPN. In contrast, mainstream one-stage detectors do not perform this standardization, resulting in each stride having unique dimensions and producing predictions independently.

Since FMap is tailored for one-stage detectors, each stride is characterized separately within its own space as each prediction emanates from only one of the strides. This means that features from different strides are processed by distinct layers, and hence may not share the same number of dimensions or the same semantic meaning. The workflow of FMap comprises the following steps:
\begin{enumerate}[leftmargin=*]
\item First FMap iterates over the in-distribution (ID) data -- namely, the training dataset -- to collect all the extracted feature maps of objects correctly predicted by the trained model, together with the stride from which each object prediction is produced.

\item Subsequently, for each class and stride, we cluster its embedding space and aggregate all object embeddings inside every cluster into a centroid, which is done by computing the mean of every feature within the cluster. This yields multiple centroids for each class in each stride, serving as the representatives of the concepts in the embedding space corresponding to that class.

\item The score for each query sample is defined as the minimum distance of the instance's features to the centroids of the predicted class in the respective stride (since each prediction originates from only one stride). Based on this definition, to determine whether a query sample is an OoD instance, we first establish a threshold for the aforementioned score. To this end, we iterate again over the ID data to obtain the scores for the correctly predicted instances. Each class in each stride will have its own ID score distribution and corresponding threshold $\lambda^{\varphi,c}$.

\item Finally, we establish the threshold value to that such that a desired True Positive Rate (TPR) is achieved for each class in each stride independently, ensuring that a specified number of ID samples are correctly classified as ID. 
\end{enumerate}

This computed threshold $\lambda^{\varphi,c}$ for every (class, stride) pair is used during inference to determine whether each prediction issued by the object detector belongs to the in-distribution or, instead, must be declared as an unknown object. The next subsection provides further mathematical details on this inference process.

\subsection{Low-level Description of FMap}\label{ssec:fmap_low_level}

This section provides a more detailed description of the FMap detector, along with a schematic summary in Algorithm~\ref{alg:fmap}. As previously mentioned, the first step is to characterize the ID by creating the representative features or centroids of each class in each stride by making use of clustering techniques. Hence, we start extracting the features from the ID samples by iterating over the training dataset $\mathcal{D}_{tr}$ and collecting the predictions. The dataset consists of images $\mathbf{x} \in \mathbb{R}^{W\times H\times C}$ with objects represented by targets $\mathbf{y} = \{\mathbf{b}, c\}$, where $\mathbf{b} \in \mathbb{R}^4$  denotes the bounding box coordinates and $c\in\{1,2,\ldots,C\}$ the class or category of the object. $C$ represents the number of classes in $\mathcal{D}_{tr}$.  Each image can contain one or multiple targets (labeled objects). Among all the predictions $\widehat{\mathbf{y}} = \{\mathbf{\widehat{b}}, \widehat{c}\}$ generated by the model, only the correctly predicted ones 
 are considered for feature extraction, i.e., $\mathcal{Y}_{\text{correct}} = \{\widehat{\mathbf{y}}\textup{ such that }\widehat{c} = c \text{ and IoU}(\widehat{\mathbf{b}}, \mathbf{b})\geq \Gamma_{\textup{IoU}}\}$, where IoU denotes \emph{Intersection over Union} and $\Gamma_{\textup{IoU}}\in\mathbb{R}(0,1)$ is a predefined threshold. Features $\mathbf{f} \in \mathbb{R}^{D_{\varphi}}$ are extracted from the feature maps $\mathbf{v}$, where ${D_{\varphi}}$ represents the dimensionality of the stride indexed by $\varphi \in \{1,2, \ldots,\zeta\}$ being $\zeta$ the total number of strides (line \textbf{5} of Algorithm \ref{alg:fmap}).


To perform the extraction of the features corresponding to each prediction, we utilize the \emph{RoIAlign} operation as described in \cite{he_mask_2017}. The inputs are a bounding box $\mathbf{b}$, the desired output size (height and width) for the features $\mathbf{f}$ of the box, the complete feature maps $\mathbf{v}$, and the ratio between the spatial resolution of the bounding boxes (height and width of the original image) and the spatial resolution of the feature maps (their height and width). The operation outputs the desired features $\mathbf{f}$ specific to the bounding box $\mathbf{b}$ out of the whole feature maps $\mathbf{v}$. It is crucial to emphasize that each prediction originates from a specific stride, thereby implying that its features possess the dimensionality of that specific stride. In particular, the number of channels are maintained after the \emph{RoIAlign} operation.

Following the process outlined in the previous subsection, we obtain a collection of features $\mathbf{f}^{\varphi, c}$ for each stride $\varphi$ and class $c$. These features are independently grouped using a clustering algorithm to yield a number $M^{\varphi, c}$ of clusters $\{\bm{\mathcal{Q}}^{m,\varphi, c}\}_{m=1}^{M^{\varphi, c}}$ (line \textbf{10} of Algorithm \ref{alg:fmap}). Features for each (class, stride) combination are clustered separately. The number of clusters per class can be established as a hyper-parameter, or instead determined by optimizing an internal clustering validation metric. Next, by using an aggregation function $f_{agg}(\cdot)$, we aggregate the features of all samples in each cluster, and obtain the centroids $\mathbf{f}_{\odot}^{m, \varphi, c}$, which can be regarded as the representative of each class $c$ and each stride $\varphi$.
\begin{algorithm}
\DontPrintSemicolon
\caption{Proposed FMap detector.}
\label{alg:fmap}
\KwData{Training dataset $\mathcal{D}_{tr}$, validation dataset $\mathcal{D}_{val}$, query (test) dataset $\mathcal{D}_{query}$, number of classes $C$, number of strides $\zeta$}
\KwResult{Detector ready to identify unknown objects}
\texttt{//Characterization of ID samples}\;
Collect the features from correct predictions in $\mathcal{D}_{tr}$:\;
\For{$\mathbf{x} \in \mathcal{D}_{tr}$}{
    Collect correct predictions $\widehat{\mathbf{y}} \in \mathcal{Y}_{\text{correct}}$\;
    Extract their features $\mathbf{f}^{\varphi, c} \in \mathbb{R}^{D_{\varphi}}$ using \emph{RoIAlign}\;
}
\textbf{Cluster} features for each class $c$ and each stride $\varphi$:\;
\For{$c \in \{1, 2, \ldots, C\}$}{
    \For{$\varphi \in \{1, 2, \ldots, \zeta\}$}{
        Obtain clusters $\{\bm{\mathcal{Q}}^{m, \varphi, c}\}_{m=1}^{M^{\varphi, c}}$ with a clustering algorithm\;
        \For{$m \in \{1, \ldots, M^{\varphi, c}\}$}{
            Compute centroids $\mathbf{f}_{\odot}^{m, \varphi, c}$ using aggregation function $f_{agg}$\;
        }
    }
}
\texttt{//Computation of the thresholds}\;
\textbf{Compute} score distribution for $\mathcal{D}_{val}$ using Eq. \eqref{eq:score_function_obj_det}:\;
\For{$\mathbf{x} \in \mathcal{D}_{val}$}{
    Compute scores $f_{OoD}(\mathbf{\widehat{y}})$ for all correct predictions\;
}
\textbf{Determine} thresholds $\lambda^{\varphi, c}$ to achieve a certain TPR over the validation set $\mathcal{D}_{val}$ with the obtained score distribution\;
\texttt{//Inference}\;
\textbf{Classify} predictions $\widehat{\mathbf{y}}$ from query samples $\mathbf{x}_{\mathit{query}}\in \mathcal{D}_{query}$ into ID or OoD using Eq. \eqref{eq:OoD_decision_obj_det}\;
\end{algorithm}

Now we can define the score function $f_{OoD}(\cdot)$ of our method as shown in Eq. \eqref{eq:score_function_obj_det}, which assigns an score to every prediction $\mathbf{\widehat{y}}$ based on the features $\mathbf{f}^{\varphi, \widehat{c}}$, calculated as the minimum distance $d(\cdot)$ between the features of the prediction and the centroids of the clusters of the predicted class and corresponding stride:
\begin{equation}\label{eq:score_function_obj_det}
    f_{OoD}(\mathbf{\widehat{y}}) = \text{min}\{d(\mathbf{f}^{\varphi, \widehat{c}}, \ \mathbf{f}_{\odot}^{m, \varphi, c})\}^{M^{\varphi, \widehat{c}}}_{m=1}.
\end{equation}

Finally, we compute the score distribution of our ID and impose a threshold $\lambda$ to determine whether a query sample is considered as ID or an OoD sample. To this end, we iterate over the ID (this time typically the validation split $\mathcal{D}_{val}$), obtaining the scores of all the correct predictions $\mathbf{\widehat{y}}$ using the above defined score function $f_{OoD}(\mathbf{\widehat{y}})$ (line \textbf{19} of Algorithm \ref{alg:fmap}). This process allows computing the threshold $\lambda^{\varphi, c}$ for each class $c$ and each stride $\varphi$ to achieve the desired TPR, i.e., the percentage of ID samples inside a validation subset $\mathcal{D}_{val}$ that will be correctly classified as ID.

At inference time, each query sample $\mathbf{x}_{query}$ in a test dataset $\mathcal{D}_{query}$ will be processed through the model, obtaining some predictions $\widehat{\mathbf{y}}$ that will be classified as ID or OoD as:
\begin{equation}\label{eq:OoD_decision_obj_det}
    \widehat{\mathbf{y}}\text{ belongs to}\begin{cases}
      \text{ID} & \text{if }f_{OoD}(\mathbf{\widehat{y}})\leq \lambda^{ \varphi, \hat{c}}, \\
      \text{OoD} & \text{if }f_{OoD}(\mathbf{\widehat{y}}) > \lambda^{ \varphi, \hat{c}}.
    \end{cases} 
\end{equation}









\subsection{Supervised Dimensionality Reduction}\label{ssec:fmap_sdr}

Since the algorithm operates on feature maps that come from the feature extraction part of an object detection model, it typically deals with high-dimensional features, namely, high values of $D_\varphi$ in the previously introduced algorith. The large dimensionality of feature maps can make feature maps become increasingly sparse due to the large volume of the feature space grows, making their distance-based characterization less effective towards detecting OoD objects. 

To alleviate this effect, we propose to apply SDR techniques to the extracted features $\textbf{f}$, so that FMap is able to distinguish between ID and OoD features more accurately. SDR techniques combine unsupervised dimensionality reduction algorithms with supervised loss functions to produce low-dimensional representations of the data, while preserving clear class boundaries. The SDR method can be represented as a model or trainable function $g$, dependent on parameters $\theta$, which maps the features space from dimensions $\mathbb{R}^{D}$ to $\mathbb{R}^{D'}$, with $D'<D$. In the case of single-stage object detectors, as the dimensionality of the features varies depending on the stride $\varphi$, we need to train a distinct function for each stride in the object detection model. Hence, the function for the SDR method can be expressed as:
\begin{equation}\label{eq:sdr}
    g_{\theta}^{\varphi}(\mathbf{f}^{\varphi}) : {\mathbb{R}^{D_{\varphi}}} \to \mathbb{R}^{{D_{\varphi}'}}.
\end{equation}

To train parameters $\theta$, a labeled dataset of features must be provided for each stride $\varphi$, where the labels correspond to the classes to which the features belong. In FMap, this labeled dataset is obtained during the feature collection from the ID data (lines \textbf{6} and \textbf{7} of Algorithm \ref{alg:fmap}).
At that stage of the FMap detector, right before creating the clusters, we can train each function $g_{\theta}^{\varphi}$ with those extracted features. Subsequently, we use them to transform the features to the new dimensionality, specifically $\mathbf{f}^{\varphi} \to \mathbf{f'}^{\varphi}$, where $\mathbf{f'}^{\varphi} \in \mathbb{R}^{{D_{\varphi}'}}$.

Once SDR has been applied over the feature maps, the FMap detector operates as described in Algorithm \ref{alg:fmap}, with the key difference that each time the features of a prediction are extracted, they are transformed to the new dimensionality using the corresponding function $g_{\theta}^{\varphi}(\mathbf{f}^{\varphi})$. This process occurs in lines \textbf{19} and \textbf{23} of Algorithm \ref{alg:fmap}.










\subsection{Enhanced Unknown Object Localization}\label{ssec:fmap_eul}





One key aspect of the usage of OoD techniques for unknown object detection is their reliance on the predictions generated by the single-stage model, which in turn depend on the inference confidence threshold, a user-selected hyperparameter. The inference confidence threshold represents the minimum confidence value a prediction must achieve to be considered valid and subsequently be output by the model. Specifically, these OoD methods can only identify an unknown object if it has been incorrectly predicted as one of the known classes, at which point the algorithm must detect this prediction as an OoD instance or unknown object. Hence, the inference confidence threshold controls the relation between the precision in detecting known objects and the recall of unknown objects: lowering its value would probably result in more unknown objects detected, but at the penalty of increasing the missed classifications of known objects.

\begin{figure*}[htb]
    \centering
    \includegraphics[width=1\textwidth]{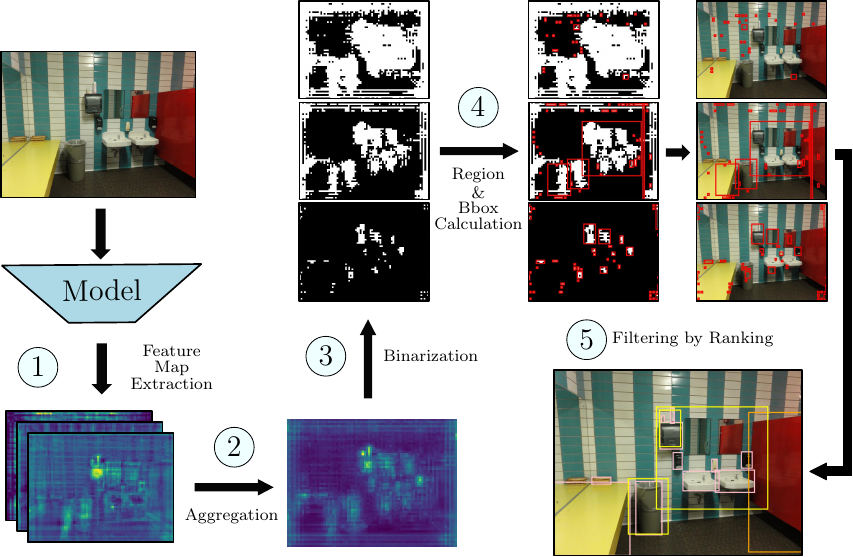}
    \captionsetup{width=1\textwidth}
    \caption{Graphical example of the application of the Enhanced Unknown Localization (EUL) algorithm. Red boxes in step \textcircled{\raisebox{-0.9pt}{4}} are bounding box proposals. In the final image after step \textcircled{\raisebox{-0.9pt}{5}}, the orange box is an unknown object detected by the basic FMap algorithm, whereas yellow are the boxes selected by EUL, and finally the pink boxes are the unknown object annotations.}
    \label{fig:unk_enhancement}
\end{figure*}

To reduce this dependency, we propose an unsupervised algorithm to generate proposals for potential unknown objects regardless of the inference confidence threshold, i.e., to improve the unknown recall of the FMap detector. This additional algorithm, hereafter referred to as \emph{Enhanced Unknown object Localization} (EUL), hinges on the hypothesis that high-resolution feature maps may contain information about unknown objects that the model does not ultimately output as predictions. The overall workflow of EUL is as follows, whose steps are illustrated in Figure \ref{fig:unk_enhancement} with numbers therein linked to the ones below:

\begin{enumerate}[leftmargin=*]
    \item[\textcircled{\raisebox{-0.9pt}{1}}] \emph{High resolution feature map extraction}: To accurately locate possible unknown objects in the image, we extract the feature maps from the stride $\varphi\in\{1,\ldots,\zeta\}$ corresponding to the highest resolution.
    \item[\textcircled{\raisebox{-0.9pt}{2}}] \emph{Feature map information aggregation}: we next condense the information from all channels of the feature maps to yield a saliency map, i.e., an image with the same resolution but only one channel, which highlights areas of the image that, in this case, likely contain objects.
    \item[\textcircled{\raisebox{-0.9pt}{3}}] \emph{Saliency map binarization}: the saliency map is converted into a binary format. This involves setting a threshold value, such that pixels with saliency values above this threshold are set to $1$ (indicating potential object presence), while those below the threshold are set to $0$ (indicating background). Multiple thresholds can be selected in order to obtain several binarized images to search for objects of different sizes. The selection of the thresholds can be done manually as a hyperparameter, or automatically tuned by resorting to methods like Otsu thresholding 
    \cite{otsu_threshold_1975}.
    \item[\textcircled{\raisebox{-0.9pt}{4}}] \emph{Region and bounding box calculation}: After applying the threshold(s), EUL utilizes connected-component labeling to identify contiguous regions, each representing a potential object in the original image. For each of these regions, we compute the bounding box surrounding the region at hand by finding the minimum and maximum coordinates within each connected component. Since the bounding boxes are calculated at the spatial resolution of a feature map, they need to be rescaled to the original image size.
    \item[\textcircled{\raisebox{-0.9pt}{5}}] \emph{Filtering by ranking}: The previous operations generate a large number of proposals; the meaningful ones are retained by using a ranking method and by selecting the top-k best proposals. To this end, features of all these proposals are first extracted through the \emph{RoIAlign} operation.
    Subsequently, for each class, the minimum distance between these features and the clusters of the class are computed. Is important to remark that the utilized stride is the one with the highest resolution, as previously discussed. Hence, each feature is associated to a vector with the minimum distance to each class. These distances are then consolidated into a single metric by computing the entropy of a vector of pseudo-probabilities built from such distances, which quantifies how far the feature vector is from the representative of all classes within the stride with largest resolution. If we denote the distances as $\{d_1,d_2,\ldots,d_C\}$, the computed entropy is given by:
    \begin{equation}
    H(\widehat{\mathbf{b}}) = -\sum\limits_{c=1}^C d_c^\ast \log_C {d}_c^\ast,
    \end{equation}
    where $d_c^\ast=d_c/\sum_{c=1}^C d_c$, and $H(\widehat{\mathbf{b}})=1$ if $d_c=1/C$ $\forall c\in\{1,\ldots,C\}$. As a result, each bounding box is ranked by a single entropy value computed on its features. Proposals exhibiting the lowest entropy are preferentially retained, based on the hypothesis that lower entropy values correlate with a higher probability of identifying a genuine object.
\end{enumerate}







\section{Experimental Setup}\label{sec:exp_setup}

In order to assess the performance of the proposed FMap detector, we design an extensive experimentation aimed to inform the responses to four research questions (RQ), formulated as follows:
\begin{itemize}[leftmargin=*]
    \item RQ1: Which is the optimal configuration for the proposed FMap OoD detector?
    \item RQ2: How does FMap perform when compared to logits-based post-hoc OoD detection methods in single-stage object detectors?
    \item RQ3: Does a fusion of feature-based methods with logits-based methods outperform other potential ensemble configurations?
    \item RQ4: How do unknown object detection algorithms implemented on single-stage models perform when compared to the state of the art?
\end{itemize}

The remainder of this section offer details on the benchmark (Subsection \ref{ssec:benchmark}, evaluation metrics (Subsection \ref{ssec:obj_det_metrics}), implementation details (Subsection \ref{ssec:impl_details}), post-hoc methods for comparison (Subsection \ref{ssec:posthoc}) and fusion strategies (Subsection \ref{ssec:fusion_strat_explanation}), all under consideration for the experiments reported in the manuscript. For the sake of replicability and to foster follow-up studies, scripts and results have been made in a public GitHub repository available at: \url{https://github.com/aitor-martinez-seras/OoD_in_Object_Detection.git}.

\subsection{Evaluation Benchmark} \label{ssec:benchmark}

The evaluation is based on the UOD benchmark presented in \cite{liang_unknown_2023}. It follows the practice of previous works on the topic of open-world object detection by using the \texttt{PASCAL VOC} dataset \cite{everingham_pascal_2010} as the training ID dataset, which contains annotations of 20 object categories. For test purposes, the UOD benchmark uses two subsets of the \texttt{COCO} dataset
\cite{lin_microsoft_2014}, namely \texttt{COCO-OoD} and \texttt{COCO-Mix}. The former contains only objects that are unknown, whereas the latter mixes both known and unknown instances. In both cases, unknowns are instances of the other 60 classes contained in \texttt{COCO}, jointly with new unknown instances labeled by \cite{liang_unknown_2023}, with images being handpicked. More details of this benchmark can be found in their paper. The number of known and unknown objects are shown in Table \ref{tab:coco_ood_coco_mixed}. This benchmark is used in the experiments of all research questions formulated above.
\begin{table}[ht]
    \centering
    \resizebox{\textwidth}{!}{\begin{tabular}{cccccc}
    \toprule
        & Dataset & \makecell[c]{Number classes\\\footnotesize{(total/known/unknown)}} & \makecell{Number of\\images} & \makecell[c]{Known\\objects} & \makecell[c]{Unknown\\objects} \\
        \midrule
        ID & \texttt{Pascal VOC} & 20/20/0 & 11,530 & 27,450 & - \\ 
        OoD & \texttt{COCO-OoD} & 60+1/0/60+1 & 504 & - & 1,655 \\ 
        OoD & \texttt{COCO-Mix} & 80+1/20/60+1 & 897 & 2,533 & 2,658 \\ 
    \bottomrule
    \end{tabular}}
    \captionsetup{width=0.9\columnwidth}
    \caption[Number of images and annotated known and unknown objects per subset in the Unknown Object Detection benchmark]{Number of images and annotated known and unknown objects per subset in the UOD benchmark. The `+1' in the \texttt{COCO-OoD} and \texttt{COCO-Mix} datasets refer to the additional class defined by the authors of \cite{liang_unknown_2023} to collectively refer to newly annotated unknown objects over the original \texttt{COCO} dataset.}
    \label{tab:coco_ood_coco_mixed}
\end{table}

\subsection{Evaluation Metrics}\label{ssec:obj_det_metrics}

To gauge the performance of the models in detecting known and unknown objects, we consider several metrics widely used in the literature related to this area \cite{liang_unknown_2023}. To evaluate the performance of known object detection we consider the mean Average Precision (mAP) considering that an object is spatially detected in the image when the IoU between a bounding box prediction and the true bounding box of the object is at least $0.5$. In order to evaluate the performance in OoD or unknown object detection, we first define $\textup{TP}_u$ as the true positive proposals of unknown objects (i.e., correctly detected unknowns); $\textup{FN}_u$ for false negative proposals (namely, unknowns detected as knowns), and $\textup{FP}_u$ for false positive proposals (corr. known objects detected as unknown). Based on these definitions, we utilize the following performance metrics:
\begin{itemize}[leftmargin=*]
    \item The Unknown Average Precision (U-AP), the average precision computed only over the unknown class.
    \item The Recall Rate (U-REC) and Precision Rate of Unknown objects (U-PRE), defined as:
    \begin{equation}
        \text{U}\texttt{-}\text{REC} = \frac{\text{TP}_u}{\text{TP}_u + \text{FN}_u},
\qquad
        \text{U}\texttt{-}\text{PRE} = \frac{\text{TP}_u}{\text{TP}_u + \text{FP}_u}.
    \end{equation}

    \item The Unknown F1-Score defined as the harmonic mean of U-PRE and U-REC, given by:
    \begin{equation}
        \text{U}\texttt{-}\text{F1} = \frac{2 \: \cdot \: \text{U}\texttt{-}\text{PRE} \: \cdot \: \text{U}\texttt{-}\text{REC}}{\text{U}\texttt{-}\text{PRE} \\ - \\ \text{U}\texttt{-}\text{REC}}.    
    \end{equation}
    
    \item The Absolute Open-Set Error (A-OSE), by \cite{miller_dropout_2018}, employed to report the count of unknown objects that are wrongly classified as any of the known classes.
    
    \item The Wilderness Impact (WI) metric, defined in \cite{dhamija_overlooked_2020} as:
    \begin{equation}
        \text{WI} = \frac{\text{Precision in closed-set}}{\text{Precision in open-set}} - 1,
    \end{equation}
    which is used also to characterize the case that unknown objects are confused with known objects.
\end{itemize}


\subsection{Implementation Details}\label{ssec:impl_details}

In terms of sotware implementation, we rely on the model templates provided by the Ultralytics library \cite{jocher_ultralytics_2023}, selecting YOLOv8 as the one-stage object detector used throughout the experiments. The number of strides $\zeta$ is 3. We trained the model from scratch over the \texttt{PASCAL VOC} ID training dataset for 300 epochs, using a batch size equal to 16. 

         

\paragraph{RoIAlign and aggregation operations}We use average as the operation to perform the aggregation of the features $f_{agg}(\cdot)$ in the vanilla FMap method. For \emph{RoIAlign}, it is necessary to choose the height and width of the resulting features. We opt for a one-pixel ($1 \times 1$) output size, hence every prediction's features will be of size equal to the number of channels in the corresponding stride.

\paragraph{Determining the thresholds for the OoD methods}FMap requires determining a threshold for each class and stride. Therefore, we need sufficient samples in each case for the thresholds to be representative of the ID data. Therefore, instead of only using the validation split from the training dataset (\texttt{PASCAL VOC} in this case), we harness all samples within the train and validation datasets to compute these thresholds. Furthermore, the TPR used to determine the thresholds $\lambda^{\varphi, c}$ is set at 95\%, which is a common choice in the literature.

\paragraph{Distance metrics}The scoring function of FMap $f_{OoD}(\cdot)$ uses a distance computation to asses the dissimilarity between features. To evaluate different options for this distance we will consider $L_1$, $L_2$ and $\mathit{cosine}$ distances for the purpose.

\paragraph{Clustering algorithms}FMap can be combined with any clustering technique, so we have tested a few in this research, namely $\mathit{KMeans}$ and $\mathit{HDBSCAN}$. To optimize them, we make a grid search of their different hyperparameters, and select the configuration that attains the best Silhouette score, individually for each class in each stride. For the case of $\mathit{KMeans}$, as the optimization algorithm tends to create 2 to 3 clusters in each case, we have also tested to force the algorithm to create 10 clusters. In addition, we include the case where all data is assumed to belong to a single, large cluster, denoting this case as $\mathit{One}$. 

\paragraph{Supervised dimensionality reduction}To implement SDR, we have chosen the \emph{IVIS} framework presented in \cite{szubert_structure-preserving_2019}. Essentially, the technique works by utilizing a Siamese network architecture trained by a variant of the standard triplet loss. After a grid search, the selected hyperparameter values are an output embedding dimension of 32 ($D'_\varphi = 32$) and a $k$ value of 15, which controls the number of nearest neighbors to retrieve for each point. The same output dimension is selected for the three strides. We refer to \cite{szubert_structure-preserving_2019} for more details of the inner workings of IVIS.

\paragraph{Enhanced unknown object localization}For the feature map aggregation, we compute the mean absolute deviation. For the binarization, a fast variation of the recursive Otsu thresholding method 
\cite{otsu_threshold_1975} is used.


\subsection{Post-hoc OoD Detection Methods} \label{ssec:posthoc}

The post-hoc OoD detection methods implemented for our experimental benchmark are selected based on their model-agnostic nature, i.e. they can be directly applicable to any type of classification branch within a neural network model, without any ambiguity of how they should implemented.


Post-hoc techniques like MSP 
\cite{hendrycks_baseline_2016}, Energy \cite{liu_energy-based_2020} and ODIN \cite{liang_enhancing_2017}, which only rely on the output of a classification branch, meet the criteria of being directly applicable to a one-stage object detector. We refer them as \emph{logits-based methods}, as they only operate on the logits of the classification branch to compute their OoD scores. These three mentioned methods are the ones included in our benchmark. In case of ODIN, the input perturbation is not applied as it is not clear how it should be implemented when multiple predictions can be issued per every image.



It is important to note that the YOLOv8 model does not produce a vector of class probabilities for each bounding box prediction. Instead, it outputs a vector of class \enquote{confidence} values, obtained by applying the sigmoid activation function in the final layer. For the purposes of this study, we have used the raw output of this final layer, prior to the application of the sigmoid activation function, as the model's internal information used by the implemented post-hoc techniques to elicit their OoD scores.

\subsection{Fusion Strategies}\label{ssec:fusion_strat_explanation} 

In response to RQ3, we investigate the performance of ensembles of feature-based (like the proposed FMap) and logits-based OoD techniques. The fusion of FMap with the considered post-hoc algorithms finds its motivation in the results of \cite{martinez2023novel}, where fusion strategies were also explored for detecting OoD instances in image classification tasks. 

\begin{figure}[bht]
    \centering
\includegraphics[width=0.7\columnwidth]{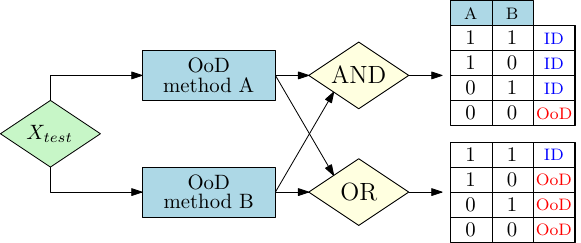}
    \vspace{-2mm}
    \captionsetup{width=\columnwidth}
    \caption[Illustration of the AND and OR fusion strategies]{Illustration of the AND and OR fusion strategies. A decision of `1` indicates that the method identifies the prediction as ID, whereas `0` indicates the contrary.}
    \label{fig:fusion_and_or}
\end{figure}

When fusing two OoD techniques, it is crucial to establish a criterion to resolve discrepancies between the predictions of both methods. One option can be to employ an AND criterion, by which a sample is declared to be OoD if both methods classifies it as such. This approach typically results in increased precision (U-PRE) but decreased recall (U-REC) for unknown objects. In contrast, an OR criterion would yield a sample classified as OoD if any of the methods (or both of them) predicts the sample as such. Consequently, an OR fusion rule potentially increases the recall at the expense of precision for both known and unknown objects. Figure \ref{fig:fusion_and_or} summarizes the operational mechanics of the AND and OR voting strategies.


With the goal of balancing precision and recall, we propose a soft voting strategy coined as SCORE. This strategy requires that each method to be fused computes a fusion score via a designated fusion function, \(f_\mathit{fusion}(\cdot)\). This score quantifies the extent to which a sample is considered ID, based on the deviation of the OoD score for a given OoD technique from the corresponding threshold. The fusion score ranges from -1 to 1, where 1 signifies a strong likelihood of the sample being ID, and \mbox{-1} indicates a strong likelihood of the sample being OoD. These scores are aggregated to produce a final fusion score that determines the sample's classification: 
\begin{equation}
    f_\mathit{fusion}(\widehat{\mathbf{y}}) = f^{A}_{OoD}(\widehat{\mathbf{y}}) + f^{B}_{OoD}(\widehat{\mathbf{y}}),
\end{equation}
\begin{equation}\label{eq:OoD_decision_fusion_obj_det}
    \widehat{\mathbf{y}}\text{ is classified as }\begin{cases}
      \text{ID} & \text{if } f_\mathit{fusion}(\widehat{\mathbf{y}}) > 0, \\
      \text{OoD} & \text{if } f_\mathit{fusion}(\mathbf{\widehat{y}}) \leq 0,
    \end{cases}
\end{equation}
where superscripts A and B represent the scores from the OoD methods A and B that are fused together.


\begin{figure*}[bht]
    \centering
    \captionsetup[subfloat]{width=0.48\textwidth}
    \subfloat[Fusion score for FMap]{%
        \includegraphics[width=0.495\textwidth]{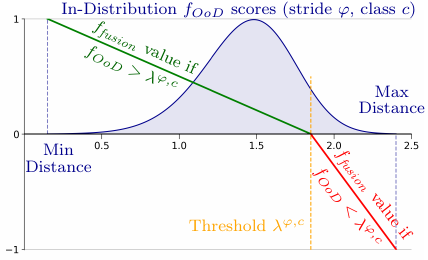}
        \label{fig:fusion_score_sub_fmap}
    } 
    \subfloat[Fusion score for logits]{%
        \includegraphics[width=0.495\textwidth]{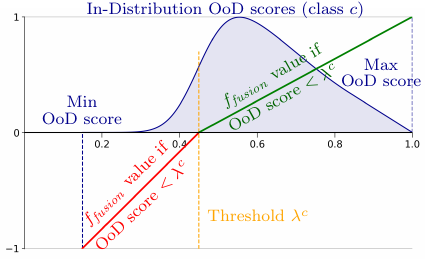}
        \label{fig:fusion_score_sub_logit}
    }
    \captionsetup{width=1\textwidth}
    \caption[Illustration of the SCORE fusion strategy]{Example of the application of the SCORE strategy. Each method computes a fusion score for $f_{\mathit{fusion}}(\cdot)$ using the piece-wise functions depicted in green and red. Values outside the defined range are clipped to the minimum and maximum values, respectively.}
    \label{fig:fusion_score}
\end{figure*}

Figure \ref{fig:fusion_score} illustrates the calculation of the fusion score for feature maps and logits-based methods. The process involves collating the minimum and maximum ID values, which subsequently define the boundaries of the piece-wise function \(f_\mathit{fusion}(\cdot)\). Depending on whether the OoD score for a particular method falls below or above the threshold, the corresponding segment of the piece-wise function is utilized to compute the fusion score.

\subsection{State-of-the-art Object Detectors} \label{ssec:baselines}

In our experiments to address RQ4, we consider several avant garde methods for object detection in the presence of unknowns, including ORE \cite{joseph_towards_2021}, VOS \cite{du_vos_2022}, OW-DETR \cite{gupta_ow-detr_2022}, UnSniffer \cite{liang_unknown_2023} and RandBox \cite{wang_random_2023}. For each of these comparison baselines, we use the hyperparameter configuration indicated in their publications. All these methods are within those included the UOD benchmark \cite{liang_unknown_2023}.

\section{Results and Analysis}\label{sec:results}


We now proceed by presenting and discussing on the experimental results obtained to address the RQs posed in the previous section. First, we aim to determine which FMap configuration performs best (RQ1, Subsection \ref{ssec:rq1_optimal_config}). Then, we compare FMap to existing post-hoc methods (RQ2, Subsection \ref{ssec:rq2_post_hoc}). Subsequently, we combine FMap with post-hoc methods to ascertain if the resulting ensemble of feature-based and logits-based methods elicits better results than when used separately (RQ3, Subsection \ref{ssec:rq3_fusion}). Finally, we compare our findings against state-of-the-art OoD detection methods designed for object detection models (RQ4, Subsection \ref{ssec:rq4_sota_comparison}).

\subsection{Preamble}\label{ssec:owod_results_preamble}

Since this paper gravitates on single-stage object detection models with open-world detection capabilities, we focus on finding the optimal configuration of the selected YOLOv8 object detector by balancing the performance in detecting both known and unknown objects. Hence, we focus on mAP as the representative score quantifying the capability of the model to detect known instances, whereas U-F1 is the indicator of the method's effectiveness in recognizing unknown objects. Specifically, for the latter we use the sum of unknown F1 (hereafter referred to as U-F1 in the text and U-F1$_{\mathit{SUM}}$ in the plots) obtained on both \texttt{COCO-OoD} and \texttt{COCO-Mix}, while mAP is calculated only from \texttt{COCO-Mix} since \texttt{COCO-OoD} includes only annotations for unknowns objects.


Building on these methodological choices, it is important to highlight the insight presented in Subsection \ref{ssec:fmap_eul}, where we emphasize the significance of the inference confidence threshold. This threshold can be regarded as a variable that calibrates the balance between the model's known and unknown object detection capabilities. Thus, during the performance assessment of the single-stage model (RQ1, RQ2 and RQ3), we aim to delineate the trade-off between known and unknown detection metrics arising when configuring the inference confidence threshold. We represent this trade-off by means of two-dimensional scatter plots, with the different configurations of cluster methods and distance metrics distinguished by unique markers and colors. For each configuration, several inference confidence thresholds are tested and plotted. The optimal configurations are those which are non-dominated by any other configuration/technique (in the Pareto sense), i.e. those for which there is no other configuration/technique that improves one objective without worsening the other. Non-dominated configurations/techniques are highlighted with red circles and connected by a red dashed line. Additionally, their A-OSE values are displayed alongside each red circle. Finally, results of all metrics of the best configurations for every research question are given in \ref{AppendixA}.

\subsection{RQ1: Which is the optimal configuration for FMap?}\label{ssec:rq1_optimal_config}

To address this first research question, we identify the non-dominated configurations for the FMap method for its naive (\emph{vanilla}) version (Algorithm \ref{alg:fmap}), and when expanded with the SDR (Subsection \ref{ssec:fmap_sdr}) and EUL (Subsection \ref{ssec:fmap_eul}) techniques. Subsequently, we compare these optimal configurations to delineate the final front of non-dominated FMap configurations.

\begin{figure}[htb]
\includegraphics[width=\columnwidth]{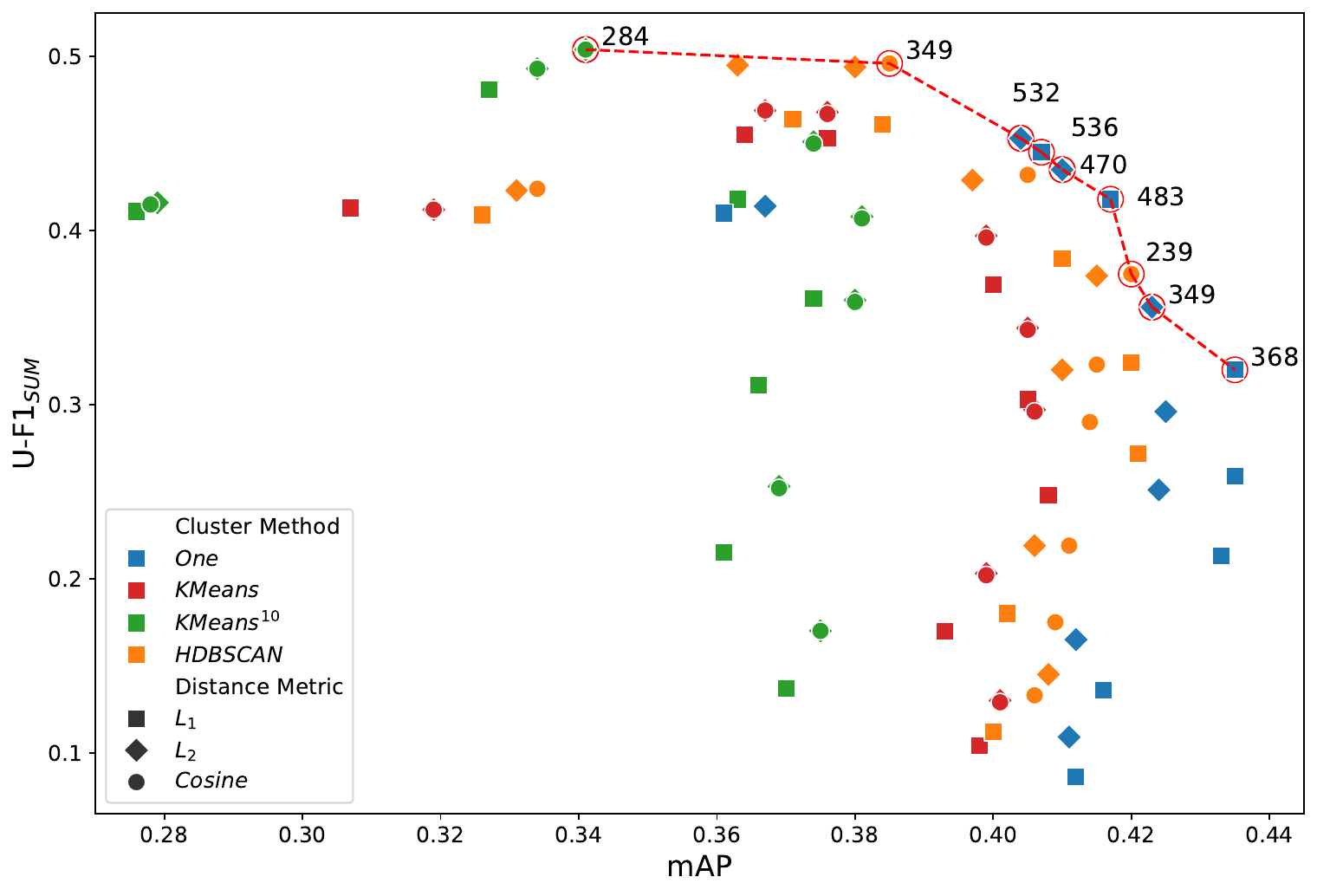}
    \captionsetup{width=\columnwidth}
    \caption[Pareto front for vanilla FMap method]{Front of non-dominated FMap configurations in the mAP versus U-F1$_{\mathit{SUM}}$ trade-off for the vanilla version of FMap. Points correspond to different configurations of the model in terms of distance metrics (represented by marker types) and clustering methods (indicated by color). For each configuration, various inference confidence thresholds are represented. We refer to Subsection \ref{ssec:owod_results_preamble} for further details.}
    \label{fig:vanilla_feature_methods}
\end{figure}

\paragraph{Vanilla FMap}
The initial analysis involves the vanilla version of the FMap method, examined in conjunction with various clustering methods and distance metrics. The comparative results are depicted in Figure \ref{fig:vanilla_feature_methods}. A first glance at this plot verifies that the choice of clustering method (indicated by color) significantly impacts on the performance of the model in detecting known objects (mAP). Generally, aggregating all features into a single large cluster per class and stride (denoted as the $\mathit{One}$ cluster method, shown in blue) results in higher mAP and lower U-F1 scores. Conversely, configurations that generate multiple clusters (thereby increasing the number of centroids for comparison) typically exhibit reduced mAP scores, but enhanced performance in detecting unknown objects. Notably, increasing the number of clusters from 2 to 3 -- which often occurs in our experiments with both $\mathit{KMeans}$ (in red) and $\mathit{HDBSCAN}$ (in orange) -- to forcing 10 clusters in $\mathit{KMeans}$ (denoted as $\mathit{KMeans}^{10}$, in green) leads to a further decrease in mAP while marginally improving U-F1. These obserbations are corroborated by examining the non-dominated configurations. Configurations yielding the highest mAP predominantly utilize the $\mathit{One}$ clustering method, while those achieving larger U-F1 scores generally employ a greater number of clusters. Notably, the best U-F1 score is achieved using  $\mathit{KMeans}^{10}$. Furthermore, with respect to the A-OSE metric (which quantifies the number of unknown instances erroneously classified as known objects) the depicted results expose a clear trend: an increase in the number of clusters correlates with a reduction in classification errors for unknown instances, i.e., lower A-OSE values.


Additionally, an analysis of the impact of distance metrics reveals significant findings. The $L_1$ and $L_2$ distance metrics typically enhance the detection of known objects, while not providing major differences between them.
Conversely, the $\mathit{Cosine}$ distance metric demonstrates superior performance in terms of robustness and the recognition of unknown objects, as evidenced by improved A-OSE and U-F1 scores, respectively.

\begin{figure}[hbt]
    \centering
    \includegraphics[width=1\columnwidth]{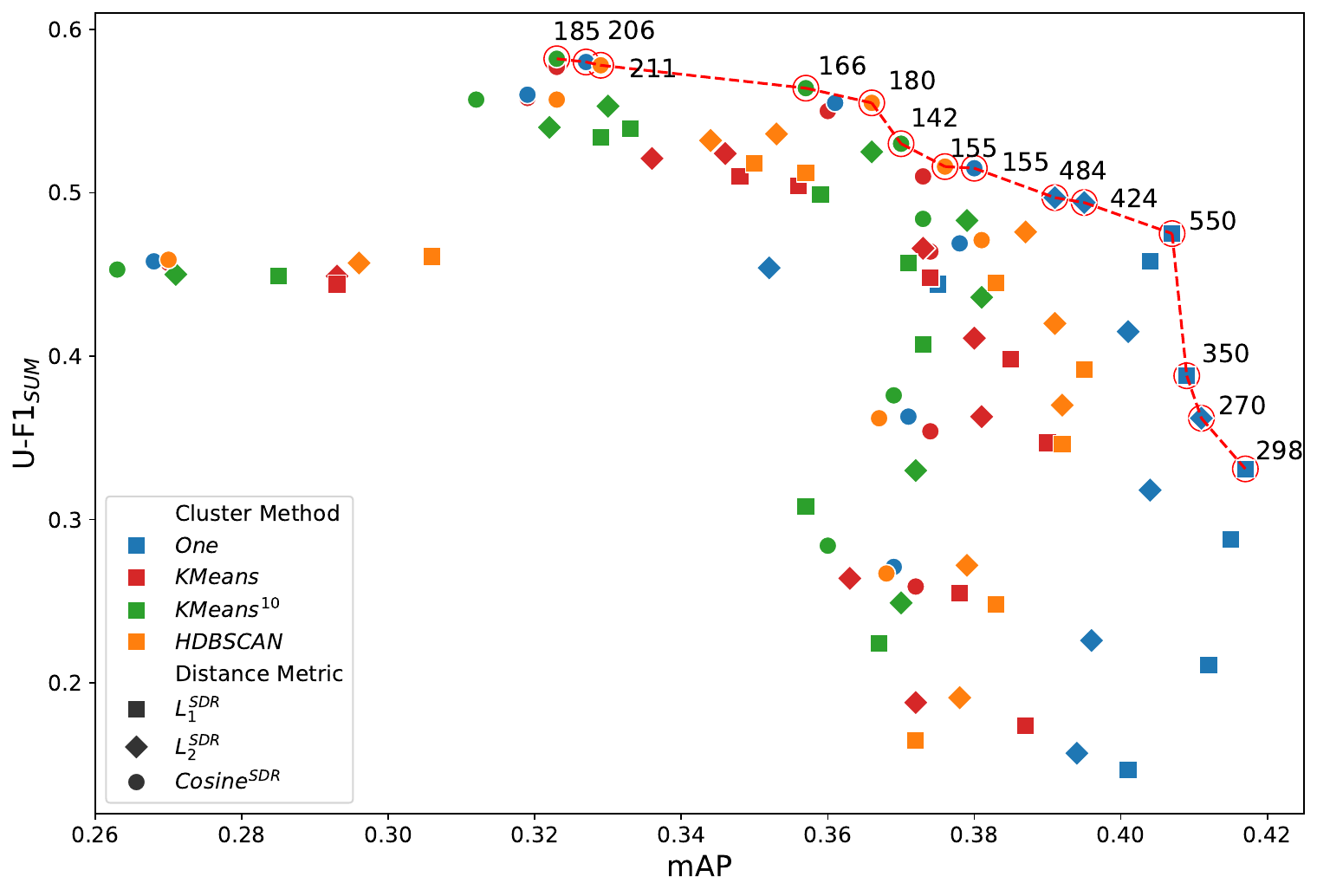}
    \captionsetup{width=\columnwidth}
    \caption[Pareto front for SDR FMap method]{Front of non-dominated FMap configurations in the mAP versus U-F1$_{\mathit{SUM}}$ trade-off for FMap with SDR. Points correspond to different configurations of the model in terms of distance metrics (represented by marker types) and clustering methods (indicated by color). For each configuration, various inference confidence thresholds are represented.}
    \label{fig:SDR_feature_methods}
\end{figure}

\paragraph{FMap with SDR}We next apply SDR to FMap to lower the dimensions of the features before their characterization via the clustering methods under consideration. Figure \ref{fig:SDR_feature_methods} summarizes the results, from which similar insights concerning the impact on performance of distance metrics and clustering methods can be drawn. In general, the $\mathit{Cosine}$ distance and a high number of clusters correlates with larger U-F1 scores. In contrast, using $L_1$ and $L_2$ metrics and only one cluster leads to increased known object performance, and slightly reduced U-F1 scores. Another key finding is that the robustness provided by the use of $\mathit{Cosine}$ distance in reducing the unknown objects wrongly classified as known classes -- lowering the \mbox{A-OSE} metric --becomes even more apparent when utilizing SDR within FMap.



\begin{figure}[htb]
    \centering
    \includegraphics[width=1\columnwidth]{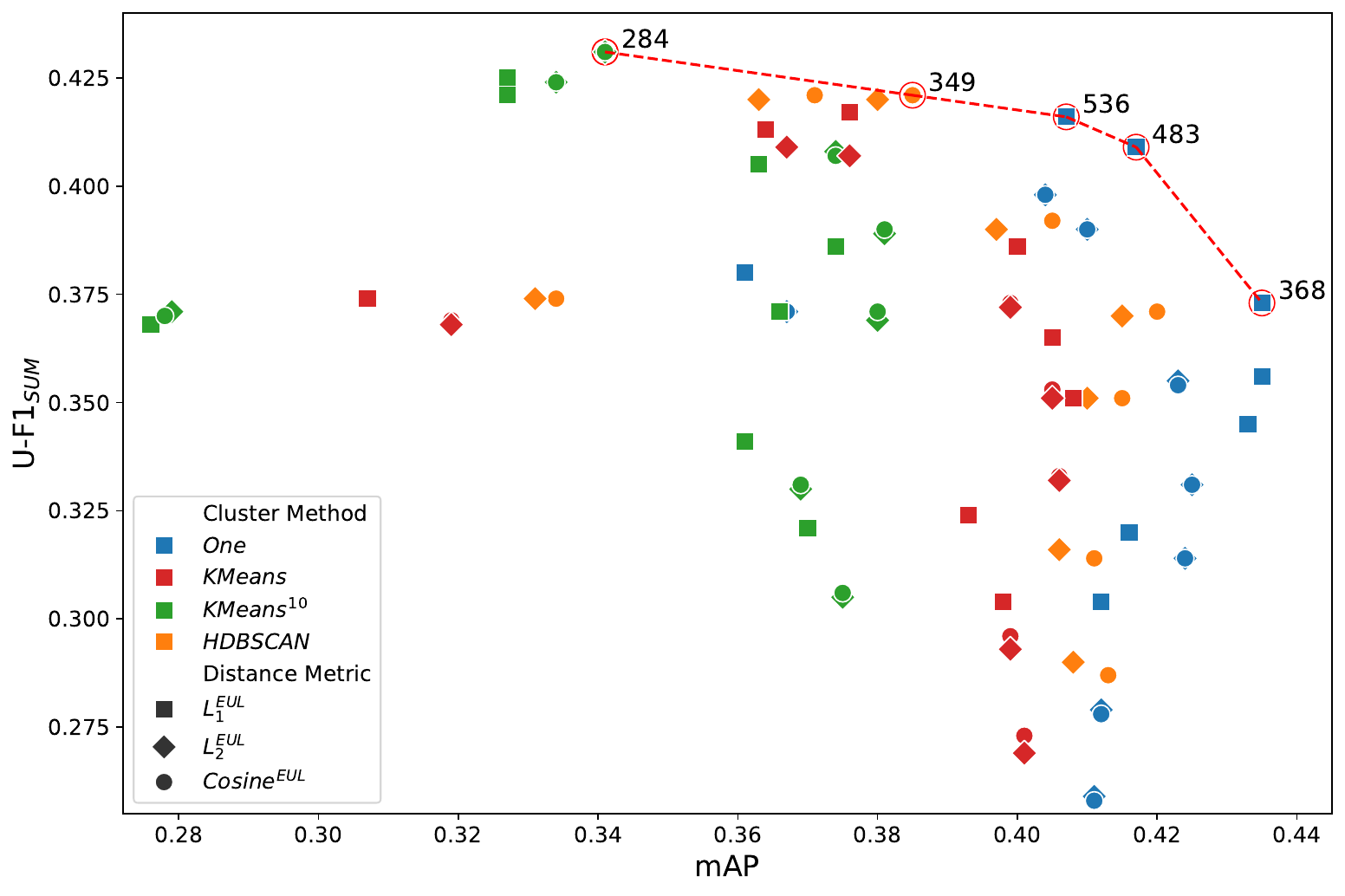}
    \captionsetup{width=\columnwidth}
    \caption[Pareto front for EUL FMap method]{Front of non-dominated FMap configurations in the mAP versus U-F1$_{\mathit{SUM}}$ trade-off for FMap with EUL. Points correspond to different configurations of the model in terms of distance metrics (represented by marker types) and clustering methods (indicated by color). For each configuration, various inference confidence thresholds are represented.}
    \label{fig:EUL_feature_methods}
\end{figure}

\paragraph{FMap with EUL}Now we apply the EUL algorithm presented in Subsection \ref{ssec:fmap_eul} to boost the unknown localization capabilities of the one-stage object detector. Consistently with earlier results, Figure \ref{fig:EUL_feature_methods} illustrates that the $\mathit{Cosine}$ distance, when paired with a clustering technique that generates more than one cluster, consistently achieves the best U-F1 scores. Meanwhile, the highest mAP is obtained using the $L_2$ metric with a single cluster. Nevertheless, in this scenario, the variability in unknown detection performance (U-F1) between the worst and best methods among the non-dominated ones is significantly smaller, narrowing from 18\% with vanilla implementation and 25\% when SDR is applied, to merely 5\% under the EUL configuration. In contrast, the performance disparity in detecting known objects (mAP) remains consistent across all three scenarios.

\begin{figure}[htb]
    \includegraphics[width=1\columnwidth]{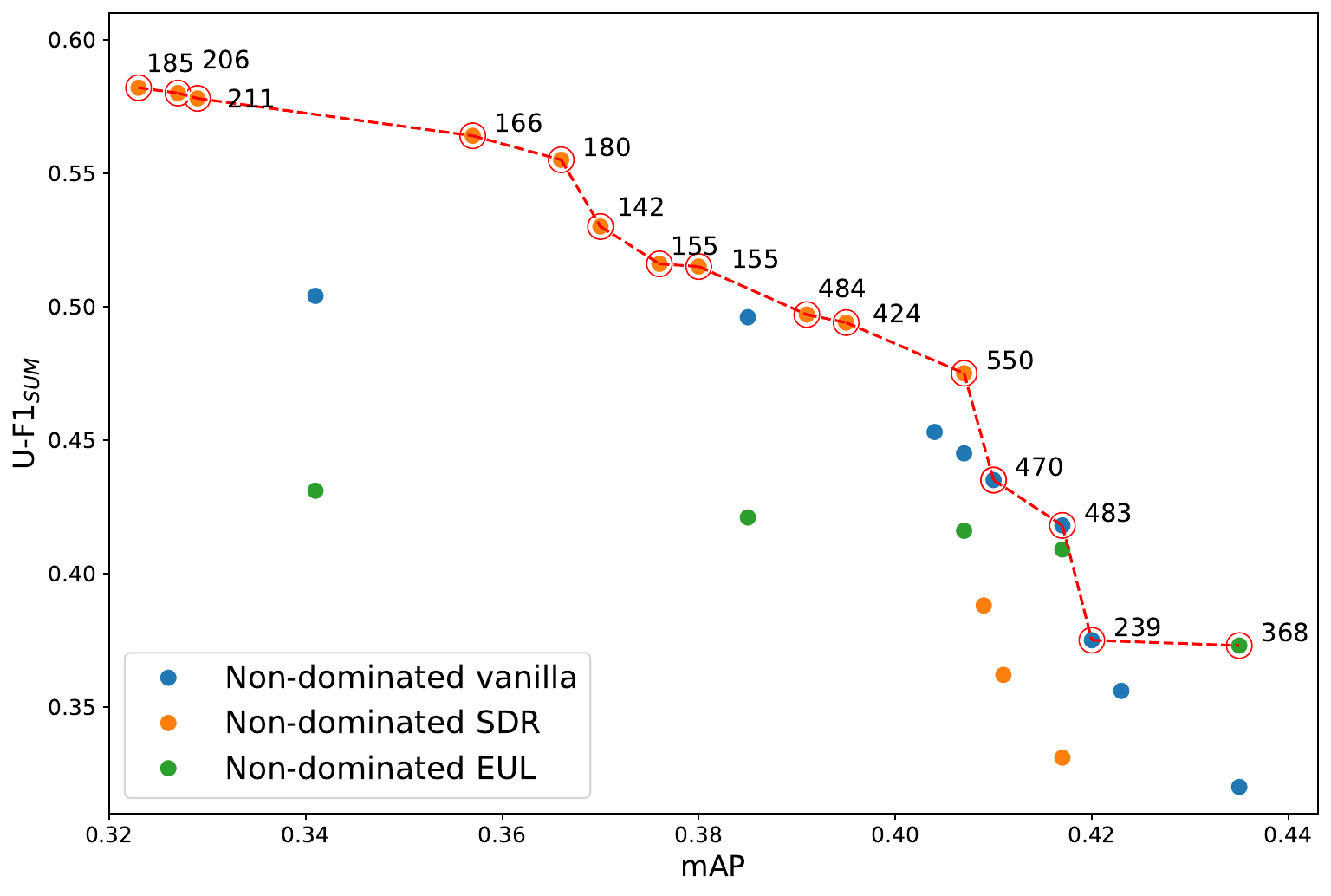}
    \captionsetup{width=\columnwidth}
    \caption[Pareto front for non-dominated FMap parameter combinations of the three versions of the algorithm (vanilla, SDR and EUL)]{mAP and U-F1$_{\mathit{SUM}}$ pareto front for non-dominated FMap parameter combinations of the three versions of the algorithm (vanilla, SDR and EUL).}
    \label{fig:ALL_feature_methods_compared}
\end{figure}

\paragraph{Comparison}Lastly, we evaluate the optimal combinations of cluster method, distance metric, and inference confidence threshold across all three versions of the algorithm (the non-dominated configurations from the vanilla FMap, FMap with SDR and FMap with EUL). This  comparative analysis is depicted in Figure \ref{fig:ALL_feature_methods_compared}. 
Results therein depicted clearly demonstrate a general trend: the FMap method enhanced with SDR consistently excels at detecting unknown objects, also achieving excellent A-OSE values when compared to other configurations.
Conversely, the vanilla FMap detector maintains the original known object detection capabilities more effectively, while delivering reasonable U-F1 scores. Finally, the addition of EUL is only beneficial when the specific combination of cluster method, distance metric and confidence threshold yields high mAP values, but low U-F1 scores. In such scenarios, EUL is capable of enhancing the unknown detection capabilities without sacrificing the mAP performance.

\begin{figure}[!htb]
    \includegraphics[width=1\columnwidth]{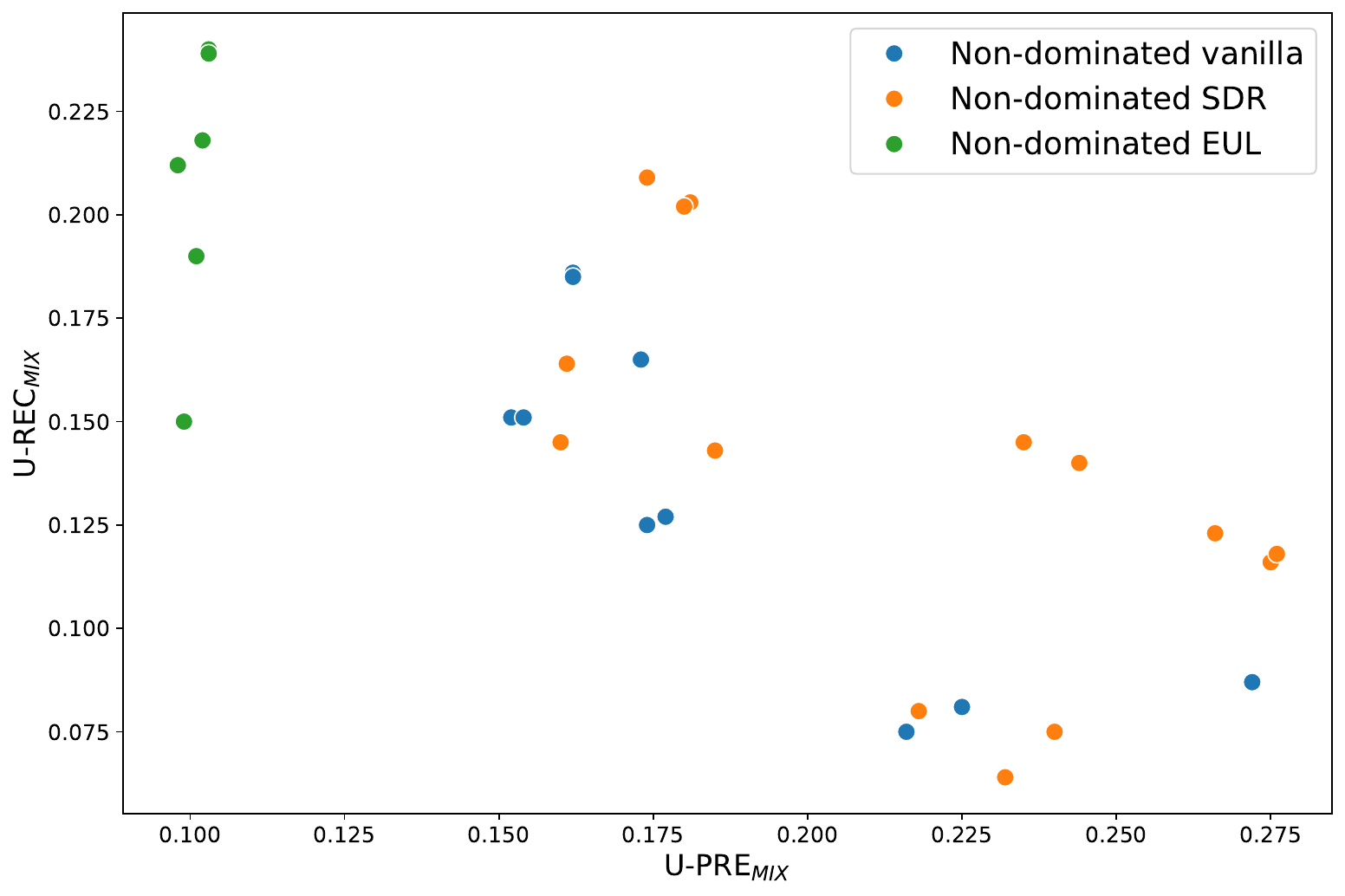}
    \captionsetup{width=\columnwidth}
    \caption[Precision vs Recall plot for non-dominated FMap versions (vanilla, SDR and EUL)]{Precision vs Recall in \texttt{COCO-Mix} subset of non-dominated FMap parameter combinations for the three versions of the algorithm (vanilla, SDR and EUL).}
    \label{fig:UNK_prec_vs_recall}
\end{figure}

Nonetheless, we note that EUL technique focuses on improving the recall of unknown objects. Its application, however, causes a decrease in precision finally leading to lower U-F1 scores. Hence, it is interesting to inspect the precision-recall trade-off in unknown object detection for these non-dominated solutions. This detailed analysis is presented in Figure \ref{fig:UNK_prec_vs_recall}, clearly revealing that EUL sacrifices precision for recall in unknown objects.

In conclusion, each version of the algorithm assessed in response to RQ1 exhibits distinct strengths and weaknesses. On the one hand, using SDR provides superior unknown detection capabilities and robustness while maintaining an acceptable known detection performance. On the other hand, EUL significantly boosts the recall of the vanilla FMap when detecting unknown objects, at the cost of considerably reducing the precision in most of the cases, hence obtaining lower U-F1 scores. Nevertheless, EUL manages to enhance the U-F1 of the vanilla FMap to a degree that it becomes one of the non-dominated parameter configurations. Finally, configurations of the vanilla FMap obtain an acceptable balance between mAP and U-F1, overriding the computational overhead associated with the other two versions under consideration. As a concluding note, in all cases $L_1$ and $L_2$ metrics perform similarly, so hereafter we will focus only on $L_1$.


\subsection{RQ2: How does FMap perform when compared to logits-based post-hoc OoD detection methods in single-stage object detectors?}
\label{ssec:rq2_post_hoc}

Our discussion follows by evaluating the performance of the optimal configurations of FMap found in the previous section, and by comparing it to that of post-hoc or logits-based techniques. These methods also depend on the inference confidence threshold set for the single-stage object detector, prompting the testing of several confidence thresholds for each post-hoc algorithm.

\begin{figure}[h!]
    \includegraphics[width=1\columnwidth]{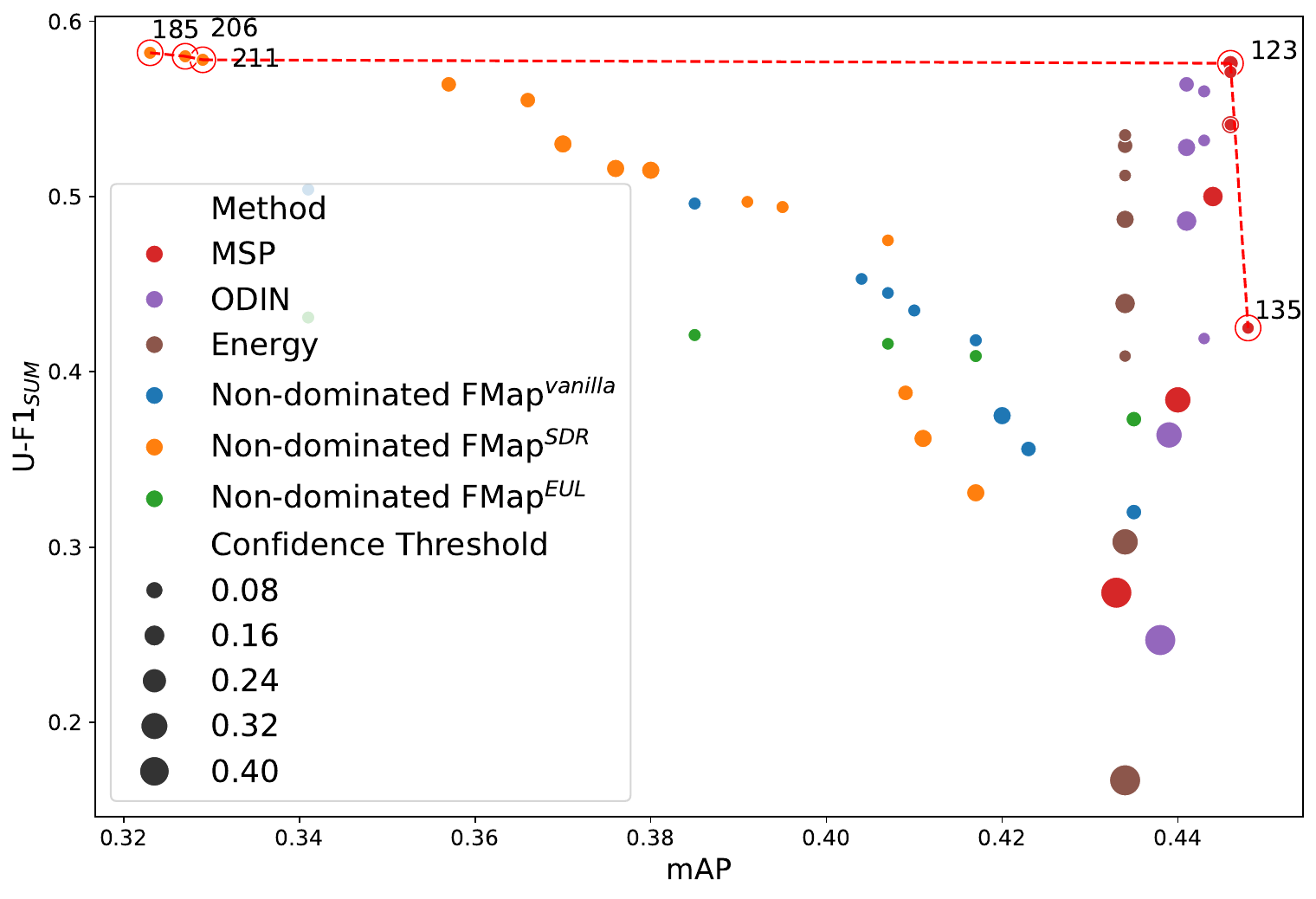}
    \captionsetup{width=\columnwidth}
    \caption[Pareto front for FMap against post-hoc methods]{Non-dominated parameter combinations of the three versions of the algorithm (vanilla FMap, FMap with SDR and FMap with EUL) against post-hoc logits-based techniques.}
    \label{fig:feature_non_dominated_vs_post_hoc}
\end{figure}

The results of this performance comparison are depicted in Figure \ref{fig:feature_non_dominated_vs_post_hoc}. Among the post-hoc methods, MSP emerges as the top performing one. When compared to the FMap method at comparable levels of unknown F1 score, post-hoc methods generally maintain better known object detection capabilities. However, the SDR version of FMap surpasses the U-F1 score of the MSP method, though this comes with a reduction in mean average precision. Summarizing, the post-hoc methods achieve similar U-F1 scores when compared to our technique while maintaining remarkable performance over the known classes. 

\subsection{RQ3: Does a fusion of feature-based methods with logits-based methods outperform other potential ensemble configurations?} \label{ssec:rq3_fusion}

The results achieved in response to RQ2 suggest an important question. Although the FMap method is marginally outperformed by post-hoc detection techniques, could it potentially serve as a superior choice for method fusion, as opposed to relying solely on the more effective logits-based methods for ensemble strategies?

\begin{figure}[htb]
    \includegraphics[width=1\columnwidth]{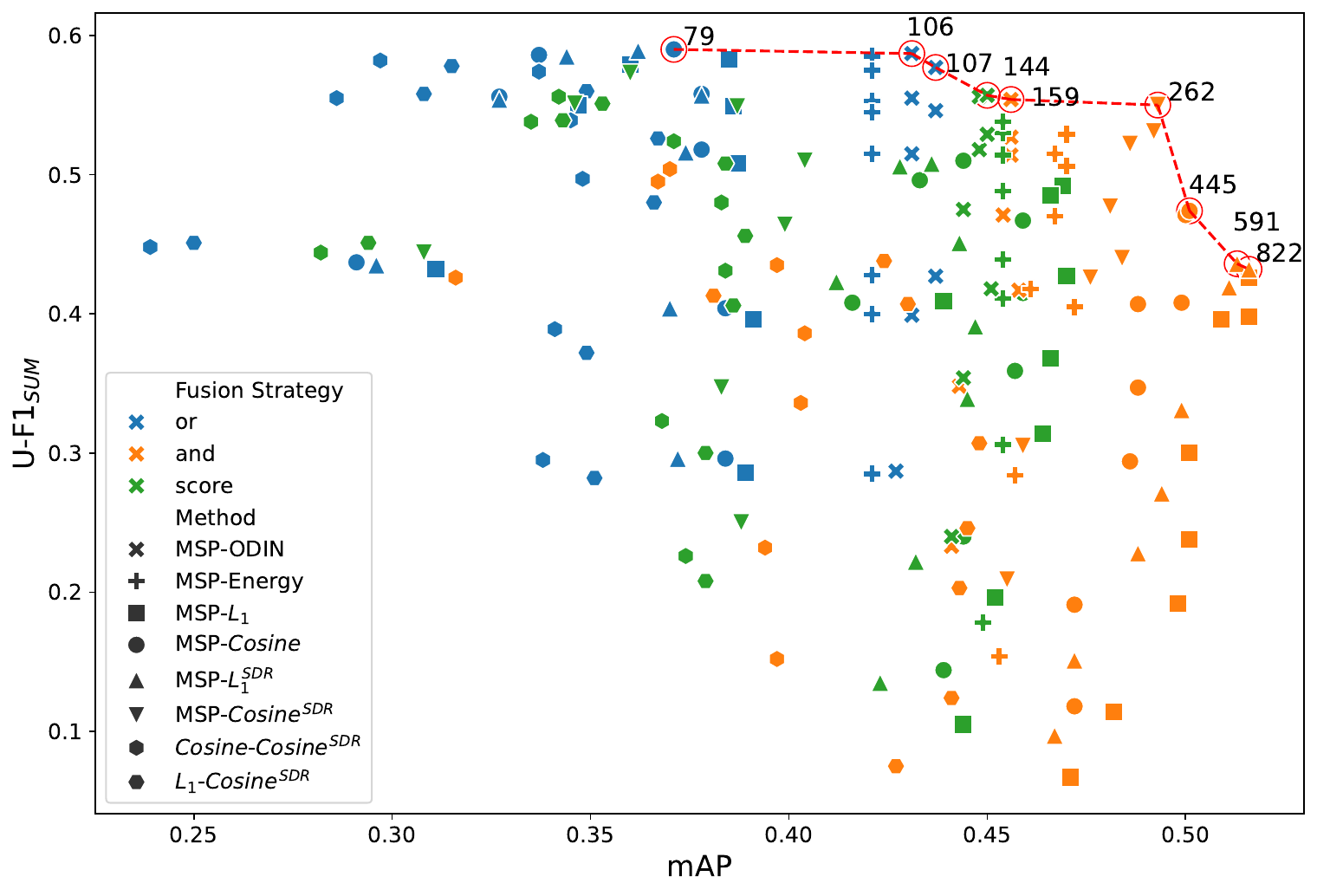}
    \captionsetup{width=\columnwidth}
    \caption[Pareto front for fusion methods]{mAP and U-F1$_{\mathit{SUM}}$ scores across various fusion possibilities of detection methods, distinguished by marker style, and for the three different fusion strategies, indicated by color. The naming convention is Method$_A$-Method$_B$, whereas the inclusion of FMap in the fusion ensemble is denoted by the distance metric in use: $L_1$, $L_1^{SDR}$, $Cosine$ and $Cosine^{SDR}$. }
    \label{fig:rq3_fusion_methods}
\end{figure}

To verify this hypothesis, which lies at the core of RQ3, we have first tested several possible ensembles together with the different fusion strategies defined in Subsection \ref{ssec:fusion_strat_explanation}. Figure \ref{fig:rq3_fusion_methods} presents the results of this prior study, where the distance metric used in the FMap method is used to name the fusion of methods. It can be clearly seen that the OR criterion (in blue) provides the best U-F1 values in exchange for known object detection capabilities, whereas the AND strategy (in orange) behaves conversely. As stated, the SCORE criterion (in green) establishes a balance in the trade-off between both metrics. Nevertheless, only one of the ensembles of this strategy is a non-dominated one. The other fusion ensemble contributing to the set of non-dominated configurations are combinations of MSP with either another post-hoc method or a feature-based technique (FMap). The fusion of different well-performing versions of FMap does not give rise to any non-dominated solution.

\begin{figure}[h!]
    \includegraphics[width=1\columnwidth]{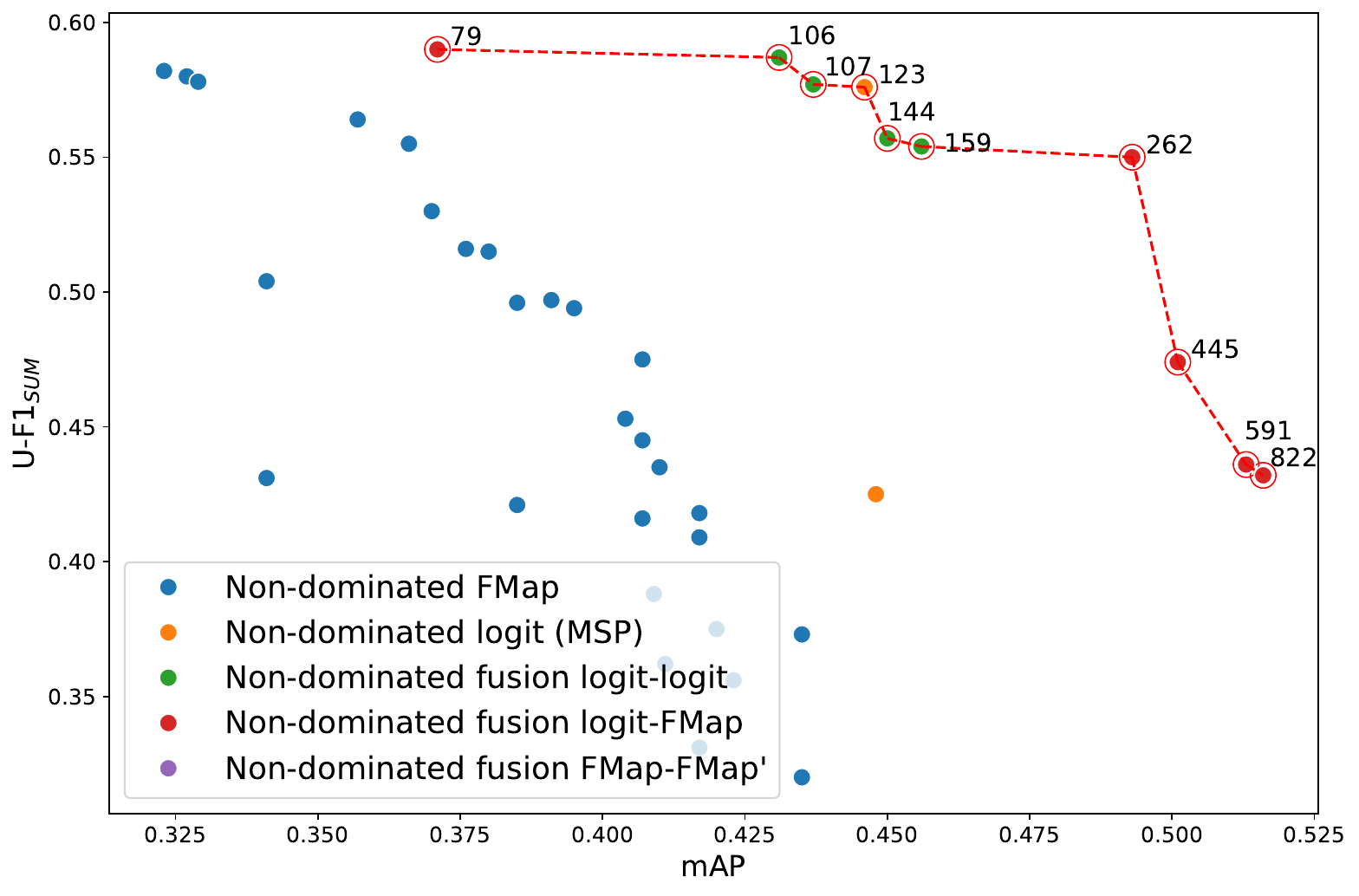}
    \vspace{-2mm}
    \captionsetup{width=1\columnwidth}
    \caption[Pareto front for best FMap, post-hoc and fusion methods]{mAP versus U-F1$_{\mathit{SUM}}$ across the best feature-based (FMap), logit-based (post-hoc) and fusion types. The fusion of different parameter combinations of FMap does not yield any non-dominated solution compared to the other ensembles, hence they do not appear in the plot. 
    }
    \label{fig:rq3_fusion_methods_against_previous}
\end{figure}

To allow for a better visualization of the final results of RQ3, we have grouped together the different ensembles attending to the type (feature- or logits-based) of the fused methods. Additionally, we compare them to the techniques previously selected in RQ1 and RQ2. This is graphically depicted in Figure \ref{fig:rq3_fusion_methods_against_previous}. We can extract valuable insights from this plot. On one hand, FMap (in blue) obtains remarkable unknown object detection capabilities for some of the configurations, but it alone can not outperform the rest the techniques when accounting for both mAP and U-F1. On the other hand, the most successful post-hoc technique, MSP, becomes part of the global set of non-dominated configurations/ensembles. In fact, it yields very similar performance to the combination of two post-hoc or logits-based methods. Finally, the combination of FMap with the best logits-based approaches (MSP) achieves the best U-F1 score for one on the possible configurations, while scoring also a superior known object detection performance in other configurations.

In conclusion, the results provided to answer RQ3 verify that the fusion two of post-hoc logit-based methods attain very similar performance in both known and unknown object detection when compared to the method alone. Conversely, fusing a feature-based method (FMap) with post-hoc detection methods ensures synergistically diverse decisions, ultimately leading to a more robust OoD detection performance. Therefore, we can confidently assure that the fusion of feature-based methods with logits-based methods outperform other fusion possibilities.
Moreover, the best fusion configurations attain better results than state-of-the-art methods, as evinced by the metrics provided in the caption of Figure \ref{fig:rq3_fusion_methods_against_previous} (not included in the plot for the sake of readability). This is further analyzed in our discussion around RQ4. 

\subsection{RQ4. How does the performance of unknown object detection algorithms implemented on single-stage models compare to that of the state of the art?} \label{ssec:rq4_sota_comparison}

We conclude our experiments with RQ4, for which we compare the best performing FMap-based methods discovered so far to the state-of-the-art methods listed in Subsection \ref{ssec:baselines}. For the sake of readability in the reported results, we have selected the most recent and well-performing comparison baselines, and we have also included the most competitive results of the vanilla, SDR and EUL versions of FMap. 
\begin{figure}[h!]
    \includegraphics[width=1\columnwidth]{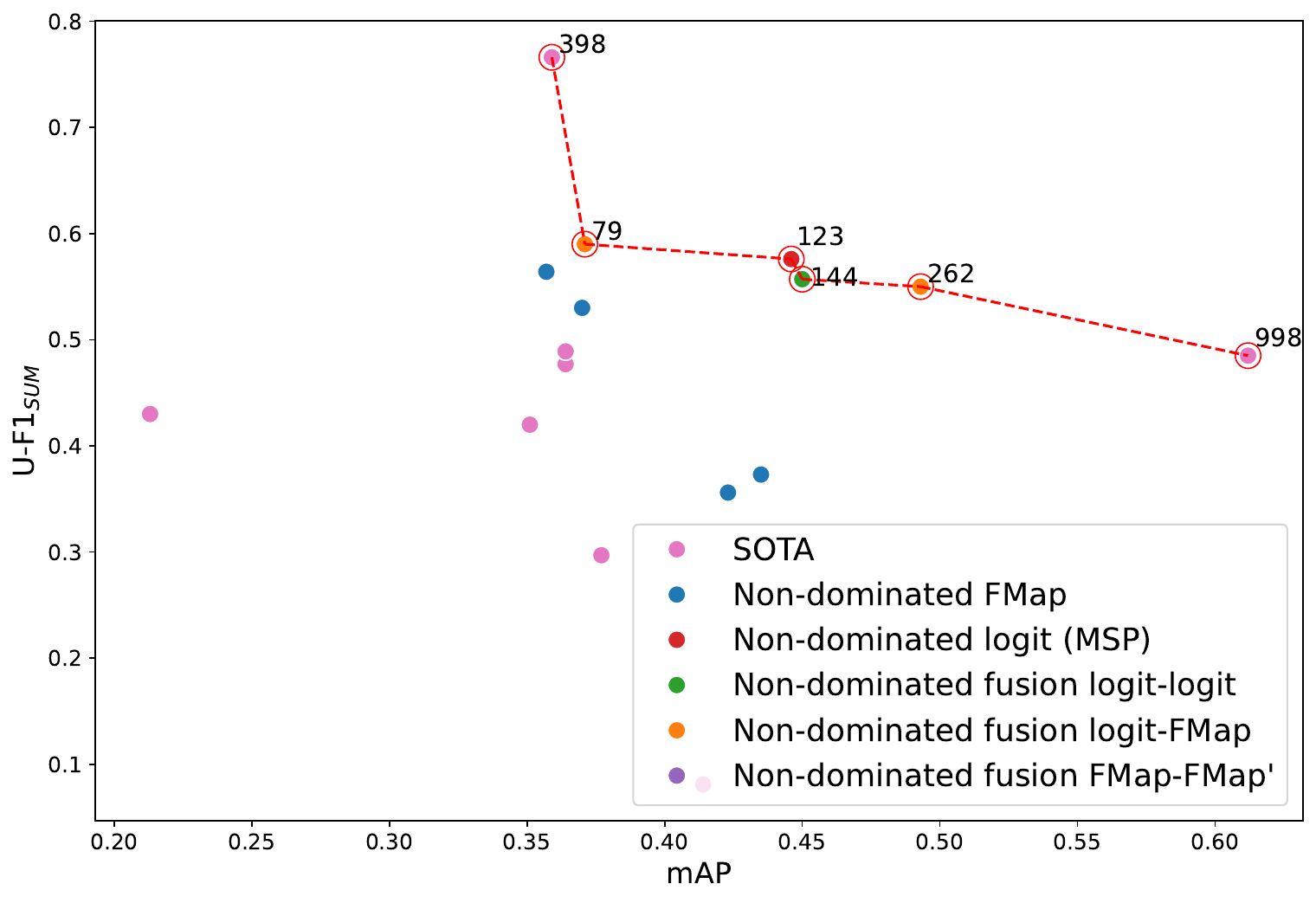}
    \vspace{-2mm}
    \captionsetup{width=1\columnwidth}
    \caption[Best non-dominated solutions in the state-of-the-art]{mAP versus U-F1$_{\mathit{SUM}}$ across the best approaches presented in this study, compared to the state-of-the-art (SOTA) methods in the OWOD literature.}
    \label{fig:rq4_SOTA_vs_ours}
\end{figure}
\begin{table*}[h]
\centering
\resizebox{\textwidth}{!}{ 
\begin{tabular}{ccclcccclccccccc}
\toprule
 &  & \multicolumn{1}{c}{\texttt{VOC}-test} & & \multicolumn{4}{c}{\texttt{COCO-OoD}} & & \multicolumn{7}{c}{\texttt{COCO-Mix}} \\
\cmidrule{3-3} \cmidrule{5-8} \cmidrule{10-16}
Method  &  Base Model  &  mAP  & & U-AP  &  U-F1  &  U-PRE  &  U-REC  & & mAP  &  U-AP  &  U-F1  &  U-PRE  &  U-REC  &  A-OSE  &  WI \\
\midrule
\multicolumn{1}{c}{\textbf{-}}  &  \multirow{9}{*}{\makecell{Faster\\R-CNN}}   & 0.483 & & -  &  -  &  -  &  -  &  & -  &  -  &  -  &  -  &  -  &  -  &  - \\
MSP \cite{hendrycks_baseline_2016} &     &  0.470  & &  0.213  &  0.314  &  0.279  &  0.359 & &  0.364  &  0.055  &  0.169  &  0.190  &  0.153  &  588  &  0.135 \\
Mahalanobis \cite{denouden2018improving} &     &  0.447 & &  0.129  &  0.271  &  0.309  &  0.241  & &  0.351  &  0.051  &  0.149  &  0.207  &  0.116  &  604  &  0.165 \\
Energy score \cite{liu_energy-based_2020} &     &  0.474 & &  0.213  &  0.308  &  0.260  &  0.377  & &  0.364  &  0.048  &  0.169  &  0.167  &  0.171  &  470  &  0.137 \\
ORE \cite{joseph_towards_2021}  &     &  0.243 & &  0.214  &  0.255  &  0.153  & \mejor 0.782 & &  0.213  &  0.140  &  0.175  &  0.103  & \mejor 0.592  &  485  &  0.089 \\
VOS$^1$ \cite{du_vos_2022} &     &  0.485 & &  0.135  &  0.196  &  0.342  &  0.137 & &  0.377  &  0.040  &  0.101  &  0.262  &  0.062  &  640  &  0.152 \\
VOS$^2$ \cite{du_vos_2022} &     &  0.469 & &  0.205  &  0.317  &  0.291  &  0.348  & &  0.364  &  0.051  &  0.172  &  0.184  &  0.163  &  409  &  0.124 \\
UnSniffer \cite{liang_unknown_2023}  &     &  0.464  & & \mejor 0.454  & \mejor 0.479  &  0.433  &  0.535 & &  0.359  & \mejor 0.150  & \mejor 0.287  &  0.222  &  0.409  &  398  &  0.175 \\
RandBox \cite{wang_random_2023}  &     &  \mejor 0.790  & & 0.266  & 0.329  &  0.243  &  0.507 & &  \mejor 0.612  & 0.052  & 0.156  &  0.112  &  0.259  &  998  &  0.109 \\


\midrule
OW-DETR \cite{gupta_ow-detr_2022}  &  \makecell[c]{Detection\\Transformer}  &  0.420  & &  0.033  &  0.056  &  0.030  &  0.380  & &  0.414  &  0.007  &  0.025  &  0.014  &  0.161  &  569  &  0.086 \\
\midrule
\multicolumn{1}{c}{\textbf{-}}  &  \multirow{9}{*}{YOLOv8}   & 0.750 & & -  &  -  &  -  &  -  & & -  &  -  &  -  &  -  &  -  &  -  &  - \\
$\mathit{Cosine}$  &    &  0.620 & &  0.160  &  0.243  &  0.655  &  0.149  & &  0.420  &  0.025  &  0.132  &  0.272  &  0.087  &  239  &  0.120 \\
$\mathit{Cosine}^{\mathit{SDR}}$  &    &  0.471 & &  0.296  &  0.392  &  0.427  &  0.363 & &  0.323  &  0.061  &  0.190  &  0.174  &  0.209  &  185  &  0.082 \\
$\mathit{Cosine}^{\mathit{SDR}}$  &    &  0.527 & &  0.238  &  0.351  & \mejor 0.668  &  0.238  & &  0.380  &  0.051  &  0.164  & \mejor 0.275  &  0.116  &  155  &  0.087 \\
$L^{EUL}_1$  &    &  0.662 & &  0.161  &  0.254  &  0.239  &  0.270  & &  0.435  &  0.031  &  0.119  &  0.099  &  0.150  &  368  &  0.141 \\
MSP  &    &  0.719 & &  0.237  &  0.390  &  0.565  &  0.298 & &  0.446  &  0.083  &  0.186  &  0.237  &  0.153  &  123  &  0.073 \\
MSP-$\mathit{Cosine}$  &    &  0.567 & &  0.289  &  0.397  &  0.563  &  0.307 & &  0.371  &  0.048  &  0.193  &  0.222  &  0.171  & \mejor 79  & \mejor 0.057 \\
MSP-$\mathit{Cosine}^{\mathit{SDR}}$  &    & 0.765  & &  0.275  &  0.362  &  0.417  &  0.320  & & 0.493  &  0.107  &  0.188  &  0.197  &  0.179  &  262  &  0.092 \\
MSP-ODIN  &    &  0.727 & &  0.271  &  0.376  &  0.397  &  0.357 & &  0.450  &  0.115  &  0.181  &  0.159  &  0.211  &  144  &  0.079 \\
\bottomrule
\end{tabular}
}
\captionsetup{width=1\textwidth}
\caption[Comparison of best FMap, post-hoc and fusion methods against the SOTA]{Comparison between state-of-the-art OWOD methods
and the best FMap-based approaches discovered in RQ1, RQ2 and RQ3. Cells corresponding to the best values for each metric in each dataset are shaded in gray. In all metrics, higher values are better except for A-OSE and WI, for which lower values are better. The hyperparameter values of the FMap-based approaches are summarized in Table \ref{tab:hyperparams_of_ours}. VOS$^1$ refers to VOS using the threshold value established in the official repository, which is calculated over the BDD100K dataset \cite{yu2020bdd100k}. VOS$^2$ refers to VOS configured with the threshold computed over the \texttt{COCO-OOD} dataset by the official code.}
\label{tab:rq4_sota_comparison}
\end{table*}

\begin{table}[!ht]
\centering
\resizebox{0.85\columnwidth}{!}{ 
\begin{tabular}{lccc}
\toprule
Method & Confidence Threshold & Cluster Method & Fusion Strategy \\
\midrule
$Cosine$ & 0.100 & HDBSCAN & - \\
$Cosine^{SDR}$ & 0.010 & KMeans$^{10}$ & - \\
$Cosine^{SDR}$ & 0.100 & ONE & - \\
$L^{EUL}_1$ & 0.050 & ONE & - \\
MSP & 0.050 & - & - \\
MSP-$Cosine$ & 0.050 & HDBSCAN & OR \\
MSP-$Cosine^{SDR}$ & 0.010 & HDBSCAN & AND \\
MSP-ODIN & 0.010 & - & SCORE \\
\bottomrule
\end{tabular}
}
\caption[Hyperparameters for our methods]{Hyperparameter values of the approaches included in Table \ref{tab:rq4_sota_comparison}.}
\label{tab:hyperparams_of_ours}
\vspace{-1mm}
\end{table}


Results are presented both in Figure \ref{fig:rq4_SOTA_vs_ours} and Table \ref{tab:rq4_sota_comparison}. Focusing on the former, we compare the above obtained non-dominated solutions for one-stage models, YOLOv8 in this case, against the state-of-the-art methods in the unknown object detection literature. It can be observed how only two of these techniques become part of the set of non-dominated solutions. UnSniffer, stands out as the method obtaining the best unknown object detection capabilities, proven by eliciting the best U-F1 score. On the other hand, RandBox achieves the best mAP of this set of solutions, while reaching acceptable U-F1. In contrast, the methodologies introduced in this study, while not achieving the highest performance across both evaluation metrics, maintain a superior equilibrium between the dual objectives. Specifically, they demonstrate strong capability for detecting unknown objects while simultaneously preserving high accuracy in identifying known entities. This is proven by the fact that three of the techniques proposed in this work are part of the non-dominated solutions in the trade-off between the performance of knowns and unknowns.

Additionally, FMap-based approaches consistently yield significantly more robust object detection models. This is evidenced by their exceptionally low A-OSE and WI metric values (see Table \ref{tab:rq4_sota_comparison}), which are the lowest among all evaluated techniques by a substantial margin—indicating a markedly reduced tendency to misclassify unknown objects as known. This robustness can be attributed to the capability of the proposed approach to eliminate the need for specialized retraining, in contrast to state-of-the-art methods that require object detectors to undergo retraining with pseudo-labels for unknown class identification. As a result, the framework facilitates the direct deployment of existing pre-trained models, thereby enhancing both practical applicability and computational efficiency.

\section{Conclusions and Future Research Lines}\label{sec:conclusions}

This work has gravitated on the detection of unknown objects in single-stage object detector models, a problem that closely relates to the open-world object detection paradigm. Specifically, we have proposed a new unknown object detection method (FMap), which characterizes the feature maps extracted by object detection models to decide whether the object predictions issued by such models correspond to known (ID) or unknown (OoD) objects. In addition, we have designed two ways of improving the performance of the vanilla FMap method. The first reduces the dimensionality of extracted features by capitalizing on existing SDR methods. The second, named EUL, aims to overcome the reduced capabilities of pretrained object detector to localize unknown objects by exploiting the information present in the feature maps computed by the model. 

Subsequently, experiments have been run over the UOD benchmark \cite{liang_unknown_2023} in response to four research questions, always bearing in mind a two-fold purpose: to detect unknown objects while maintaining a good detection on known classes. Different configurations of FMap in terms of distance metric, clustering method, the addition of SDR/EUL, and inference confidence threshold have been evaluated in terms of the trade-off between the aforementioned goals. 

The main conclusions drawn from the experiments can be summarized as follows:
\begin{itemize}[leftmargin=*]
\item Our experiments for RQ1 confirm that each version of the algorithm presented has its own unique advantages and disadvantages. The SDR method, for instance, excels in detecting unknown objects (high U-F1 values) and in providing robustness (measured by A-OSE), while preserving a satisfactory performance for known objects (measured by mAP). Conversely, integrating the EUL algorithm with FMap significantly enhances the detection of unknown objects, although this often results in a notable decrease in precision across many instances. The vanilla version obtained an acceptable balance between both objectives -- known and unknown detection -- without the additionally complexity associated with the other two.

\item Regarding RQ2, we have compared the best configurations of FMap against several post-hoc methods. The results showed that post-hoc methods generally lead to better known object detection capabilities for the same unknown object detection performance.

\item To address RQ3, we have explored different fusion strategies between the considered OoD detection methods. For the fusion, we have developed three different possible strategies to elucidate the final outcome when participating methods disagree. Experiments evince that the ensemble of two post-hoc or logits-based techniques barely improve the results obtained by each of the methods independently. In contrast, when the fusion is between the proposed FMap and a post-hoc technique, a remarkable performance boost is achieved, outperforming the methods within the ensemble when applied separately. This demonstrates that logits- and feature-based technique produce diverse yet complementary decisions, leading to a more robust detection of unknown objects.

\item Finally, RQ4 has been answered by comparing some of the best configurations found in the previous experiments against state-of-the-art methods for unknown object detection and OWOD. Results have shown that FMap-based methods for single-stage object detectors consistently dominates the benchmark over known classes. Regarding the detection of unknowns, they have yielded superior recall statistics (ORE \cite{joseph_towards_2021}) and F1 (UnSniffer \cite{liang_unknown_2023}) scores, whereas one of FMap configurations have obtained the best precision. Furthermore, FMap-based methods have attained lower A-OSE and WI values, suggesting a reduced likelihood of misclassifying unknown objects as known. This robustness stems from our strategy of avoiding retraining, dramatically reducing the rate of object misclassifications.
\end{itemize}

To conclude, in this work we demonstrate how single-stage object detectors are inherently robust and how they can be easily endowed with decent unknown object detection capabilities compared to the SOTA. Moreover, we have designed an OoD or unknown object detection method for single-stage object detectors that obtains acceptable results and is very suitable for the fusion with the existing post-hoc methods.

In conclusion, this work has highlighted the inherent robustness of single-stage object detectors and their potential to be effectively enhanced with strong unknown object detection capabilities, standing competitively against state-of-the-art methods. Additionally, we have introduced a novel approach for unknown object detection, tailored for single-stage detectors, which achie\-ves solid results when compared to existing post-hoc OoD methods. Our findings establish a new direction for further research and practical applications in open-world scenarios demanding safe and trustworthy AI deployments.

\paragraph{Future research lines}In future research, we aim to expand and refine the FMap method introduced in this manuscript. Its current implementation uses a straightforward algorithm to leverage the model’s extracted features. We envision enhancing this algorithm by exploring techniques to integrate features across different strides into a unified representation. Additionally, we plan to implement other SDR algorithms to further improve performance. Another promising avenue lies in the exploration of neural activations, particularly the activations derived from feature extraction components like feature maps. These feature maps have shown considerable potential in OoD detection, whether used independently or combined with logits-based methods. We foresee manifold opportunities in this area (e.g. adversarial refinement of activations using latent interpolation), all recognizing their value for improving model robustness without requiring model retraining.


\section*{Acknowledgments}

A. Martinez Seras receives funding support from the Basque Government through its BIKAINTEK PhD support program. J. Del Ser acknowledges funding support from the same institution through the Consolidated Research Group MATHMODE (IT1456-22) and the ELKARTEK program (BEREZ-IA, grant no. KK-2023/00012).

\newpage
\appendix

\section[Results on Unknown Object Detection Benchmark]{Results on Unknown Object Detection Benchmark} 
\label{AppendixA}

This appendix contains a table with the numerical results of all FMap configurations tested in our experiments.

\begin{table}[h]
\centering
\adjustbox{width=1.02\columnwidth, angle=90}{
\begin{tabular}{cccccccccccccccccc}
\toprule
Group & Method & \makecell[c]{Confidence\\threshold} & \makecell[c]{Distance\\metric} & \makecell[c]{Clustering\\method} & \makecell[c]{Fusion\\strategy} & mAP & U-AP & U-F1 & U-PRE & U-REC & mAP & U-AP & U-F1 & U-PRE & U-REC & A-OSE & WI \\
\midrule
RQ1 & FMap & 0.050 & $L_1$ & ONE & - & 0.662 & 0.152 & 0.209 & 0.578 & 0.127 & 0.435 & 0.022 & 0.111 & 0.216 & 0.075 & 368 & 0.141 \\
RQ1 & FMap & 0.010 & $L_1$ & ONE & - & 0.637 & 0.157 & 0.272 & 0.440 & 0.197 & 0.417 & 0.039 & 0.146 & 0.174 & 0.125 & 483 & 0.145 \\
RQ1 & FMap & 0.005 & $L_1$ & ONE & - & 0.629 & 0.200 & 0.293 & 0.381 & 0.238 & 0.407 & 0.039 & 0.152 & 0.152 & 0.151 & 536 & 0.142 \\
RQ1 & FMap & 0.050 & $L_2$ & ONE & - & 0.658 & 0.158 & 0.237 & 0.592 & 0.148 & 0.423 & 0.023 & 0.119 & 0.225 & 0.081 & 349 & 0.139 \\
RQ1 & FMap & 0.010 & $L_2$ & ONE & - & 0.640 & 0.204 & 0.287 & 0.451 & 0.211 & 0.410 & 0.040 & 0.148 & 0.177 & 0.127 & 470 & 0.141 \\
RQ1 & FMap & 0.005 & $L_2$ & ONE & - & 0.631 & 0.209 & 0.301 & 0.386 & 0.246 & 0.404 & 0.039 & 0.152 & 0.154 & 0.151 & 532 & 0.136 \\
RQ1 & FMap & 0.010 & $L_2$ & Kmeans$^{10}$ & - & 0.529 & 0.225 & 0.331 & 0.412 & 0.277 & 0.341 & 0.042 & 0.173 & 0.162 & 0.186 & 284 & 0.118 \\
RQ1 & FMap & 0.050 & $\mathit{Cosine}$ & ONE & - & 0.658 & 0.158 & 0.237 & 0.592 & 0.148 & 0.423 & 0.023 & 0.119 & 0.225 & 0.081 & 349 & 0.139 \\
RQ1 & FMap & 0.010 & $\mathit{Cosine}$ & ONE & - & 0.640 & 0.204 & 0.287 & 0.451 & 0.211 & 0.410 & 0.040 & 0.148 & 0.177 & 0.127 & 470 & 0.141 \\
RQ1 & FMap & 0.005 & $\mathit{Cosine}$ & ONE & - & 0.631 & 0.209 & 0.301 & 0.386 & 0.246 & 0.404 & 0.039 & 0.152 & 0.154 & 0.151 & 532 & 0.136 \\
RQ1 & FMap & 0.010 & $\mathit{Cosine}$ & Kmeans$^{10}$ & - & 0.529 & 0.225 & 0.331 & 0.412 & 0.276 & 0.341 & 0.042 & 0.173 & 0.162 & 0.185 & 285 & 0.118 \\
RQ1 & FMap & 0.100 & $\mathit{Cosine}$ & HDBSCAN & - & 0.620 & 0.160 & 0.243 & 0.655 & 0.149 & 0.420 & 0.025 & 0.132 & 0.272 & 0.087 & 239 & 0.120 \\
RQ1 & FMap & 0.010 & $\mathit{Cosine}$ & HDBSCAN & - & 0.577 & 0.219 & 0.327 & 0.431 & 0.264 & 0.385 & 0.045 & 0.169 & 0.173 & 0.165 & 349 & 0.125 \\
RQ1 & FMap & 0.100 & $L^{\mathit{SDR}}_1$ & ONE & - & 0.636 & 0.165 & 0.231 & 0.732 & 0.137 & 0.417 & 0.021 & 0.100 & 0.232 & 0.064 & 298 & 0.135 \\
RQ1 & FMap & 0.050 & $L^{\mathit{SDR}}_1$ & ONE & - & 0.635 & 0.167 & 0.271 & 0.647 & 0.171 & 0.409 & 0.021 & 0.117 & 0.218 & 0.080 & 350 & 0.144 \\
RQ1 & FMap & 0.005 & $L^{\mathit{SDR}}_1$ & ONE & - & 0.612 & 0.221 & 0.323 & 0.424 & 0.261 & 0.407 & 0.038 & 0.152 & 0.160 & 0.145 & 550 & 0.134 \\
RQ1 & FMap & 0.100 & $L^{\mathit{SDR}}_2$ & ONE & - & 0.597 & 0.164 & 0.248 & 0.713 & 0.150 & 0.411 & 0.022 & 0.114 & 0.240 & 0.075 & 270 & 0.127 \\
RQ1 & FMap & 0.010 & $L^{\mathit{SDR}}_2$ & ONE & - & 0.581 & 0.223 & 0.333 & 0.480 & 0.255 & 0.395 & 0.040 & 0.161 & 0.185 & 0.143 & 424 & 0.133 \\
RQ1 & FMap & 0.005 & $L^{\mathit{SDR}}_2$ & ONE & - & 0.573 & 0.224 & 0.334 & 0.408 & 0.283 & 0.391 & 0.041 & 0.163 & 0.161 & 0.164 & 484 & 0.129 \\
RQ1 & FMap & 0.100 & $\mathit{Cosine}^{\mathit{SDR}}$ & ONE & - & 0.527 & 0.238 & 0.351 & 0.668 & 0.238 & 0.380 & 0.051 & 0.164 & 0.275 & 0.116 & 155 & 0.087 \\
RQ1 & FMap & 0.010 & $\mathit{Cosine}^{\mathit{SDR}}$ & ONE & - & 0.477 & 0.292 & 0.389 & 0.433 & 0.352 & 0.327 & 0.062 & 0.191 & 0.181 & 0.203 & 206 & 0.086 \\
RQ1 & FMap & 0.100 & $\mathit{Cosine}^{\mathit{SDR}}$ & Kmeans$^{10}$ & - & 0.520 & 0.241 & 0.362 & 0.663 & 0.249 & 0.370 & 0.049 & 0.168 & 0.266 & 0.123 & 142 & 0.082 \\
RQ1 & FMap & 0.050 & $\mathit{Cosine}^{\mathit{SDR}}$ & Kmeans$^{10}$ & - & 0.507 & 0.241 & 0.384 & 0.581 & 0.287 & 0.357 & 0.047 & 0.180 & 0.235 & 0.145 & 166 & 0.085 \\
RQ1 & FMap & 0.010 & $\mathit{Cosine}^{\mathit{SDR}}$ & Kmeans$^{10}$ & - & 0.471 & 0.296 & 0.392 & 0.427 & 0.363 & 0.323 & 0.061 & 0.190 & 0.174 & 0.209 & 185 & 0.082 \\
RQ1 & FMap & 0.100 & $\mathit{Cosine}^{\mathit{SDR}}$ & HDBSCAN & - & 0.533 & 0.241 & 0.351 & 0.667 & 0.238 & 0.376 & 0.051 & 0.165 & 0.276 & 0.118 & 155 & 0.089 \\
RQ1 & FMap & 0.050 & $\mathit{Cosine}^{\mathit{SDR}}$ & HDBSCAN & - & 0.519 & 0.241 & 0.377 & 0.587 & 0.278 & 0.366 & 0.049 & 0.178 & 0.244 & 0.140 & 180 & 0.092 \\
RQ1 & FMap & 0.010 & $\mathit{Cosine}^{\mathit{SDR}}$ & HDBSCAN & - & 0.478 & 0.295 & 0.388 & 0.432 & 0.352 & 0.329 & 0.062 & 0.190 & 0.180 & 0.202 & 211 & 0.091 \\
RQ1 & FMap & 0.050 & $L^{EUL}_1$ & ONE & - & 0.662 & 0.161 & 0.254 & 0.239 & 0.270 & 0.435 & 0.031 & 0.119 & 0.099 & 0.150 & 368 & 0.141 \\
RQ1 & FMap & 0.010 & $L^{EUL}_1$ & ONE & - & 0.637 & 0.203 & 0.277 & 0.241 & 0.327 & 0.417 & 0.031 & 0.132 & 0.101 & 0.190 & 483 & 0.145 \\
RQ1 & FMap & 0.005 & $L^{EUL}_1$ & ONE & - & 0.629 & 0.206 & 0.282 & 0.233 & 0.358 & 0.407 & 0.040 & 0.134 & 0.098 & 0.212 & 536 & 0.142 \\
RQ1 & FMap & 0.010 & $L^{EUL}_2$ & Kmeans$^{10}$ & - & 0.529 & 0.225 & 0.287 & 0.234 & 0.371 & 0.341 & 0.051 & 0.144 & 0.103 & 0.240 & 284 & 0.118 \\
RQ1 & FMap & 0.010 & $\mathit{Cosine}^{EUL}$ & Kmeans$^{10}$ & - & 0.529 & 0.225 & 0.287 & 0.234 & 0.371 & 0.341 & 0.049 & 0.144 & 0.103 & 0.239 & 285 & 0.118 \\
RQ1 & FMap & 0.010 & $\mathit{Cosine}^{EUL}$ & HDBSCAN & - & 0.577 & 0.217 & 0.282 & 0.234 & 0.356 & 0.385 & 0.046 & 0.139 & 0.102 & 0.218 & 349 & 0.125 \\
\midrule
RQ2 & MSP & 0.050 & - & - & - & 0.719 & 0.237 & 0.390 & 0.565 & 0.298 & 0.446 & 0.083 & 0.186 & 0.237 & 0.153 & 123 & 0.073 \\
RQ2 & MSP & 0.001 & - & - & - & 0.724 & 0.319 & 0.289 & 0.206 & 0.483 & 0.448 & 0.107 & 0.136 & 0.086 & 0.329 & 135 & 0.078 \\
\midrule
RQ3 & MSP-FMap & 0.050 & $\mathit{Cosine}$ & HDBSCAN & OR & 0.567 & 0.289 & 0.397 & 0.563 & 0.307 & 0.371 & 0.048 & 0.193 & 0.222 & 0.171 & 79 & 0.057 \\
RQ3 & MSP-FMap & 0.005 & $\mathit{Cosine}$ & HDBSCAN & AND & 0.775 & 0.209 & 0.314 & 0.354 & 0.282 & 0.501 & 0.113 & 0.160 & 0.152 & 0.169 & 445 & 0.120 \\
RQ3 & MSP-FMap & 0.005 & $L^{\mathit{SDR}}_1$ & ONE & AND & 0.788 & 0.202 & 0.287 & 0.416 & 0.220 & 0.513 & 0.112 & 0.149 & 0.178 & 0.128 & 591 & 0.126 \\
RQ3 & MSP-FMap & 0.001 & $L^{\mathit{SDR}}_1$ & ONE & AND & 0.786 & 0.207 & 0.282 & 0.276 & 0.288 & 0.516 & 0.112 & 0.150 & 0.125 & 0.188 & 822 & 0.113 \\
RQ3 & MSP-FMap & 0.010 & $\mathit{Cosine}^{\mathit{SDR}}$ & HDBSCAN & AND & 0.765 & 0.275 & 0.362 & 0.417 & 0.320 & 0.493 & 0.107 & 0.188 & 0.197 & 0.179 & 262 & 0.092 \\
RQ3 & MSP-ODIN & 0.050 & - & - & OR & 0.702 & 0.289 & 0.399 & 0.567 & 0.308 & 0.431 & 0.077 & 0.188 & 0.228 & 0.160 & 106 & 0.069 \\
RQ3 & MSP-ODIN & 0.010 & - & - & OR & 0.702 & 0.289 & 0.393 & 0.405 & 0.381 & 0.437 & 0.094 & 0.184 & 0.157 & 0.223 & 107 & 0.069 \\
RQ3 & MSP-ODIN & 0.010 & - & - & AND & 0.734 & 0.270 & 0.375 & 0.397 & 0.355 & 0.456 & 0.109 & 0.179 & 0.158 & 0.206 & 159 & 0.085 \\
RQ3 & MSP-ODIN & 0.010 & - & - & SCORE & 0.727 & 0.271 & 0.376 & 0.397 & 0.357 & 0.450 & 0.115 & 0.181 & 0.159 & 0.211 & 144 & 0.079 \\
\bottomrule
\end{tabular}

}
\caption[Results from best method of RQ1, RQ2 and RQ3]{Results from best configurations of the methods from RQ1, RQ2 and RQ3

}
\label{tab:appA_rq1_rq2_rq3}
\end{table}

\newpage
\bibliographystyle{IEEEtran}
\bibliography{bibliography}
\vfill

\end{document}